\documentclass[runningheads]{llncs}

\usepackage{eccv}

\usepackage{eccvabbrv}

\usepackage{graphicx}
\usepackage{booktabs}
\usepackage{multirow}
\usepackage{pifont}
\usepackage{tikz}
\usetikzlibrary{arrows.meta, positioning, calc, fit, backgrounds, decorations.pathreplacing, shapes.geometric, shapes}
\newcommand{\B}[1]{{\fontseries{b}\selectfont #1\kern0pt}}
\newcommand{\cmark}{\ding{51}} 
 
\definecolor{clrRed}{HTML}{D94040}
\definecolor{clrRedBg}{HTML}{FFF0F0}
\definecolor{clrGray}{HTML}{666666}
\definecolor{clrGrayBg}{HTML}{F7F7FA}
\definecolor{clrBlue}{HTML}{3060B0}
\definecolor{clrBlueBg}{HTML}{EEF2FA}
\definecolor{clrGreen}{HTML}{2E7D32}
\definecolor{clrOrange}{HTML}{D07020}
\definecolor{clrPurple}{HTML}{7B1FA2}
\definecolor{clrOrangeBg}{HTML}{FFF8F0}
\definecolor{clrYellow}{HTML}{B08A20}
\definecolor{clrYellowBg}{HTML}{FFFDE8}
\definecolor{clrMagenta}{HTML}{E07098}
\definecolor{clrPro}{HTML}{2E7D32}
\definecolor{clrCon}{HTML}{C62828}

\usepackage[accsupp]{axessibility}  

\usepackage{hyperref}

\usepackage{orcidlink}
\usepackage[export]{adjustbox} 
\usepackage{float}
\usepackage[hypcap=false]{caption}

\newcommand{\concat}{\mathbin{;}}
\begin{document}

\title{GeoMix: Descriptor-Free Visual Localization via Global Context and Multi-Detector Training} 

\titlerunning{GeoMix}

\author{Yejun Zhang\inst{1}\orcidlink{0009-0005-2838-9518} \and
Xinjue Wang\inst{1}\orcidlink{0000-0001-7660-2444} \and
Zihan Wang\inst{1}\orcidlink{0009-0002-0695-5839} \and
Esa Rahtu\inst{2}\orcidlink{0000-0001-8767-0864} \and
Juho Kannala\inst{1,3}\orcidlink{0000-0001-5088-4041}}

\authorrunning{Y.~Zhang et al.}

\institute{Aalto University, Espoo, Finland\\
\email{\{yejun.zhang,xinjue.wang,zihan.1.wang,juho.kannala\}@aalto.fi} \and
Tampere University, Tampere, Finland\\
\email{esa.rahtu@tuni.fi} \and
University of Oulu, Oulu, Finland}

\maketitle

\begin{center}
    \resizebox{\textwidth}{!}{\begin{tikzpicture}[
    node distance=0.5cm and 0.45cm,
    box/.style={
        draw=black!40, rounded corners=3pt, fill=white,
        minimum height=0.9cm, minimum width=1.7cm,
        align=center, font=\small, inner sep=3pt, line width=0.6pt
    },
    hlbox/.style={
        draw=#1, rounded corners=3pt, fill=#1!8,
        minimum height=0.9cm, minimum width=1.7cm,
        align=center, font=\small, inner sep=3pt, line width=1.0pt
    },
    annot/.style={
        font=\scriptsize, align=center, text width=2.2cm
    },
    proscons/.style={
        font=\scriptsize, align=left, anchor=west, text width=3.6cm
    },
    arr/.style={->, >=Stealth, line width=0.7pt, color=black!55},
    bgstyle/.style={
        draw, rounded corners=5pt, inner sep=5pt
    },
    main label/.style={
        font=\small\bfseries, anchor=north west, xshift=4pt, yshift=-4pt
    },
    sub label/.style={
        font=\footnotesize\bfseries, anchor=north west, xshift=3pt, yshift=-3pt
    }
]

\node[box] (a-img) {Query\\Image};
\node[box, right=of a-img] (a-kp) {Keypoint\\Detection};
\node[box, right=of a-kp] (a-desc) {Descriptor\\Extraction};
\node[box, right=of a-desc] (a-match) {2D-3D\\Matching};
\node[box, right=of a-match] (a-pose) {6-DoF\\Pose};

\draw[arr] (a-img) -- (a-kp);
\draw[arr] (a-kp) -- (a-desc);
\draw[arr] (a-desc) -- (a-match);
\draw[arr] (a-match) -- (a-pose);

\node[proscons, right=0.5cm of a-pose] (a-pc) {
    \textcolor{clrPro}{\checkmark~High accuracy}\\[1pt]
    \textcolor{clrCon}{\texttimes~Large storage}\\[1pt]
    \textcolor{clrCon}{\texttimes~Privacy risk}\\[1pt]
    \textcolor{clrCon}{\texttimes~Map maintenance}
};

\node[box, below=1.8cm of a-img] (b-img) {Query\\Image};
\node[box, right=of b-img] (b-kp) {Keypoint\\Detection};
\node[box, right=of b-kp] (b-geo) {Geometric\\Cues};
\node[box, right=of b-geo] (b-match) {2D-3D\\Matching};
\node[box] (b-pose) at (a-pose |- b-img) {6-DoF\\Pose};

\draw[arr] (b-img) -- (b-kp);
\draw[arr] (b-kp) -- (b-geo);
\draw[arr] (b-geo) -- (b-match);
\draw[arr] (b-match) -- (b-pose);

\node[proscons] (b-pc) at (a-pc.west |- b-pose) {
    \textcolor{clrCon}{\texttimes~Low accuracy}\\[1pt]
    \textcolor{clrPro}{\checkmark~Compact storage}\\[1pt]
    \textcolor{clrPro}{\checkmark~Privacy preserving}\\[1pt]
    \textcolor{clrPro}{\checkmark~Easy maintenance}
};

\node[box, below=1.25cm of b-img] (c-img) {Query\\Image};
\node[box, right=of c-img] (c-kp) {Keypoint\\Detection};
\node[hlbox=clrBlue, right=of c-kp] (c-geo) {Geometric\\Cues};
\node[hlbox=clrOrange, right=of c-geo] (c-match) {2D-3D\\Matching};
\node[box] (c-pose) at (a-pose |- c-img) {6-DoF\\Pose};

\draw[arr] (c-img) -- (c-kp);
\draw[arr] (c-kp) -- (c-geo);
\draw[arr] (c-geo) -- (c-match);
\draw[arr] (c-match) -- (c-pose);

\node[annot, below=0.25cm of c-geo, text width=4.0cm] (ann-geo) {
    \textcolor{clrBlue}{\textbf{Local:} Dir.+Dist.} ~\textcolor{clrOrange}{\textbf{Global:} GC. Nodes}
};
\draw[clrBlue, line width=0.6pt, -{Stealth[length=4pt]}] (ann-geo.north) -- (c-geo.south);

\node[annot, below=0.25cm of c-match, text=clrMagenta] (ann-mix) {
    \textbf{Strategy:}\\Mix-Training
};
\draw[clrOrange, line width=0.6pt, -{Stealth[length=4pt]}] (ann-mix.north) -- (c-match.south);

\node[proscons] (c-pc) at (a-pc.west |- c-pose) {
    \textcolor{clrOrange}{$\uparrow$~Improved accuracy}\\[1pt]
    \textcolor{clrPro}{\checkmark~Compact storage}\\[1pt]
    \textcolor{clrPro}{\checkmark~Privacy preserving}\\[1pt]
    \textcolor{clrPro}{\checkmark~Easy maintenance}
};

\begin{scope}[on background layer]
    \coordinate (L) at ([xshift=-10pt]a-img.west);
    \coordinate (R) at ([xshift=10pt]a-pc.east);

    \path (a-img.north) ++(0, 18pt) coordinate (a-top);
    \path (a-pc.south) ++(0, 6pt) coordinate (a-bot);
    \node[bgstyle, inner sep=8pt, fill=clrRedBg, draw=clrRed!30, line width=0.6pt,
          fit=(L |- a-top) (R |- a-bot)] (bg-a) {};
    \node[main label, text=clrRed] at (bg-a.north west) {(a) Descriptor-Based Methods};

    \path (b-img.north) ++(0, 28pt) coordinate (b-top);
    \node[bgstyle, inner sep=8pt, fill=clrGrayBg, draw=black!25, line width=0.6pt,
          fit=(L |- b-top) (R |- ann-mix.south)] (bg-b) {};
    \node[main label, text=clrGray] at (bg-b.north west) {(b) Descriptor-Free Methods};

    \path (b-img.north) ++(0, 1.4em) coordinate (b1-top);
    \node[bgstyle, fill=clrYellowBg, draw=clrYellow!50, line width=0.6pt,
          fit=(b-img) (b-pose) (b-pc) (b1-top)] (bg-b1) {};
    \node[sub label, text=clrYellow!80!black] at (bg-b1.north west) {i. Prior Works};

    \path (c-img.north) ++(0, 1.4em) coordinate (c-top);
    \path (ann-mix.south) ++(0, 4pt) coordinate (c-bot);
    \path (ann-geo.south)  ++(0, 4pt) coordinate (c-bot-geo);
    \node[bgstyle, fill=clrBlueBg!50, draw=clrBlue!60, line width=1.0pt,
          fit=(c-img) (c-top) (c-bot) (c-bot-geo) (c-pose) (c-pc)] (bg-c) {};
    \node[sub label, text=clrBlue] at (bg-c.north west) {ii. GeoMix (Ours)};
\end{scope}

\end{tikzpicture}}%
    
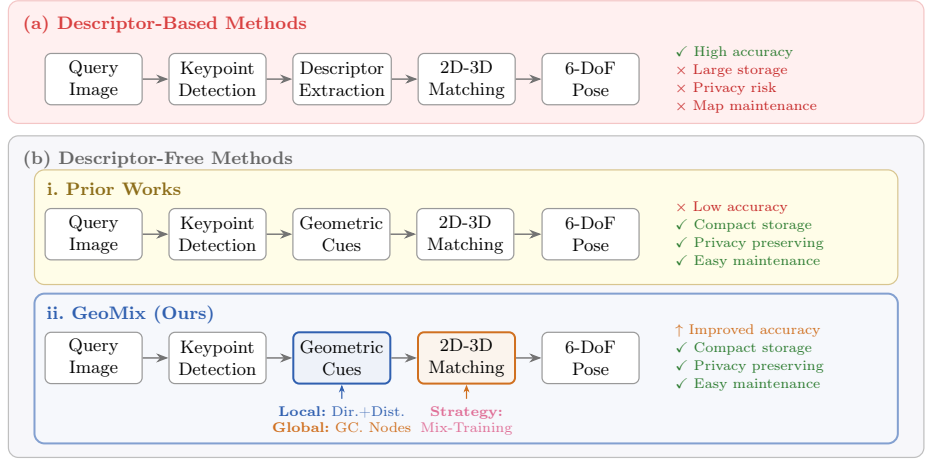
\captionof{figure}{Comparison of visual localization pipelines. (a)~Descriptor-based methods achieve high accuracy but require large storage, raise privacy concerns, and incur costly map maintenance. (b-i)~Descriptor-free methods offer compact, privacy-preserving localization yet lag in accuracy. (b-ii)~GeoMix bridges this gap with directional and distance-aware geometric embeddings, global context nodes, and a Mix-Training strategy. Dir.: direction; Dist.: distance; GC: global context.}
    \label{fig:teaser}
\end{center}

\begin{abstract}
Descriptor-free visual localization eliminates high-dimensional descriptor storage, preserves scene privacy, and simplifies map maintenance, yet its accuracy still lags far behind descriptor-based pipelines. We identify this gap to insufficient geometric discriminability in geometry-only matching.
Without visual appearance, current methods underutilize local geometry cues, lack the global context among keypoints, and overfit to a single keypoint detector. 
We further observe that descriptor-free matching naturally enables multi-detector training, as heterogeneous keypoints can be optimized in a shared geometry-only space without aligning descriptor spaces.
Building on these insights, we propose GeoMix, a descriptor-free 2D-3D matching framework that strengthens geometric discriminability at three levels. Locally, directional and distance-aware embeddings enrich neighborhood aggregation with fine-grained spatial structure. Globally, learnable context nodes aggregate and redistribute scene-wide information via cross-attention to resolve ambiguities beyond local receptive fields. 
At the training level, Mix-Training exploits this detector-agnostic geometry space to learn representations across multiple keypoint detectors.
Extensive experiments on MegaDepth, Cambridge Landmarks, 7Scenes, and Aachen Day-Night show that GeoMix sets a new state of the art among descriptor-free methods, reducing 75th-percentile rotation error by 89\% and translation error by up to 90\% over the previous best, while generalizing zero-shot to unseen detectors and narrowing the gap to descriptor-based pipelines. Code is available at \url{https://github.com/YejunZhang/Geomix}.

\end{abstract}

\section{Introduction}
\label{sec:intro}

Accurate camera localization underpins autonomous navigation, augmented reality, and large-scale mapping~\cite{billinghurst2015survey, cadena2016past, fuentes2015visual, mur2017orb}. The dominant paradigm achieves high accuracy by matching local image features to a pre-built 3D map via visual descriptors and recovering pose with a PnP-RANSAC solver~\cite{sarlin2019coarse, sattler2011fast, sattler2012improving, sattler2017large, zuliani2009ransac}.

Yet this success comes at a cost: storing high-dimensional descriptors for millions of 3D points consumes substantial memory, descriptors can be inverted to reveal sensitive scene content~\cite{chelani2021privacy, pan2023privacy}, and map updates require recomputing descriptors whenever new images are observed~\cite{dusmanu2021cross, wang2024dgc, zhou2022geometry}. These constraints motivate descriptor-free localization~\cite{campbell2020solving, wang2024dgc, zhou2022geometry}, which represents keypoints through geometric cues such as keypoint positions and spatial relations, enabling compact and privacy-preserving maps. Despite these practical advantages, descriptor-free methods still lag significantly behind descriptor-based counterparts in accuracy, preventing adoption in precision-critical systems.

The accuracy gap reflects a fundamental challenge. Without visual appearance, the network must extract sufficient discriminative signal from geometry alone, and current methods~\cite{zhou2022geometry, wang2024dgc, zhang2025a2gnnangleannulargnnvisual} fall short at every stage of this process. At the local level, neighborhood aggregation in graph neural networks (GNNs) relies on basic spatial relations, overlooking directional and metric cues that could sharpen per-point discriminability. At the global scale, local graph convolution operators have limited receptive fields; repetitive structures such as corridors or building facades are geometrically indistinguishable locally and therefore produce ambiguous correspondences. At the training level, existing methods train with a single keypoint detector, causing the network to absorb detector-specific patterns rather than learning geometry that transfers across conditions.

We observe, however, that the descriptor-free formulation itself conceals an untapped advantage. In descriptor-based pipelines, each keypoint detector is tightly coupled with its own descriptor space, so switching detectors requires reconciling incompatible representations~\cite{dusmanu2021cross, jaenal2025towards, cubero2025matcha}. The descriptor-free setting removes this coupling entirely: since matching relies purely on geometric relations, keypoints from \emph{any} detector are interchangeable. This detector-agnostic property has not been exploited by prior methods, yet it opens a natural path to stronger generalization---training over diverse detectors simultaneously to force the network to learn geometry that transcends any single detector's characteristics.

Building on this observation, we propose \textbf{GeoMix}, a descriptor-free 2D--3D matching framework that strengthens \textbf{Geo}metric discriminability at all three levels through enriched local--global representations and \textbf{Mix}-Training over multiple detectors. At the local level, directional and distance-aware embeddings enrich neighborhood aggregation with fine-grained spatial structure. At the global level, learnable global context nodes aggregate and redistribute global information via cross-attention, enabling each feature to resolve ambiguities beyond its local receptive field. At the training level, our Mix-Training strategy jointly optimizes over multiple keypoint detectors, leveraging the detector-agnostic property unique to the descriptor-free setting to learn robust geometric representations.
Extensive experiments on MegaDepth, Cambridge Landmarks, 7Scenes, and Aachen Day-Night show that GeoMix sets a new state of the art among descriptor-free methods, reducing 75th-percentile rotation error by 89\% and translation error by up to 90\% over the previous best, substantially narrowing the gap to descriptor-based pipelines.

Our contributions are as follows:
\begin{itemize}
    \item We introduce GeoMix, a descriptor-free 2D--3D matching framework that strengthens geometric discriminability at both local and global levels: directional and distance-aware embeddings sharpen per-point discrimination in neighborhood aggregation, while learnable global context nodes aggregate and redistribute scene-wide information via cross-attention to disambiguate repetitive structures.
    \item We propose Mix-Training, a multi-detector training strategy uniquely enabled by the descriptor-free setting: since keypoints from any detector are interchangeable without descriptors, joint optimization forces the network to learn geometry that generalizes to entirely unseen detectors at test time.
    \item Extensive experiments across four benchmarks show that GeoMix reduces 75th-percentile rotation error by 89\% and translation error by up to 90\% over the best prior descriptor-free method, while generalizing zero-shot to unseen detectors.
\end{itemize}

\section{Related work}
\label{sec:related work}

\paragraph{Structure-Based Localization.}
Structure-based visual localization achieves high accuracy by establishing 2D--3D correspondences against a pre-built 3D map. Classical methods~\cite{sattler2011fast,sattler2012improving} relied on Structure-from-Motion (SfM) models to match features between query images and 3D points. Learning-based advances have significantly improved image retrieval~\cite{arandjelovic2016netvlad,ge2020self,gordo2017end,revaud2019learning}, with recent works~\cite{keetha2023anyloc,Izquierdo_CVPR_2024_SALAD} leveraging powerful pretrained models to further boost retrieval accuracy. Feature extraction methods~\cite{detone2018superpoint,revaud2019r2d2,dusmanu2019d2,tyszkiewicz2020disk} extract robust visual representations for reliable keypoint detection and description. For keypoint matching, SuperGlue~\cite{sarlin2020superglue} employs graph neural networks to refine correspondences, while recent advances~\cite{shi2022clustergnn,lindenberger2023lightglue,jiang2024omniglue,zhang2025comatcher} continue to push the boundaries of matching accuracy and efficiency. Detector-free methods~\cite{sun2021loftr,chen2022aspanformer,wang2024efficient} further improve accuracy by matching dense image pairs directly, though at higher computational cost. More recently, end-to-end regression methods such as Reloc3r~\cite{dong2025reloc3r} directly predict relative camera poses from image pairs, offering fast inference but requiring large-scale training data and substantial model capacity.

\paragraph{Scene Compression.}
Scene compression reduces storage via map pruning~\cite{camposeco2019hybrid,cao2014minimal,dymczyk2015gist,yang2022scenesqueezer,li2010location,laskar2024differentiable}, descriptor compression~\cite{jegou2010product,sattler2015hyperpoints,ke2004pca,dong2023learning,wang2024maddr,yang2022scenesqueezer,laskar2024differentiable}, or hybrid approaches~\cite{camposeco2019hybrid,yang2022scenesqueezer,laskar2024differentiable}. Scene coordinate regression~\cite{brachmann2021visual,brachmann2023accelerated,li2020hierarchical,wang2024hscnet++,wang2021continual,wang2024glace} learns compact representations but struggles to generalize to new scenes. While these techniques save storage, they do not fully address privacy concerns or descriptor maintenance. Our descriptor-free method is complementary and can be combined with scene compression to further reduce model size.

\paragraph{Privacy-Preserving Visual Localization.}
As demonstrated in~\cite{pittaluga2019revealing}, image details can be recovered from descriptors, raising data privacy concerns.
To address this, privacy-preserving methods~\cite{speciale2019privacy1,speciale2019privacy2,shibuya2020privacy,pan2023privacy,moon2024efficient,moon2024sphere} obscure spatial details by lifting 2D/3D points to geometric primitives such as lines, ray clouds~\cite{moon2024efficient}, and sphere clouds~\cite{moon2024sphere}, but generally suffer from accuracy degradation or require additional sensor inputs.
Other methods adopt alternative strategies, including scene landmarks~\cite{do2022learning}, semantic segmentation~\cite{pietrantoni2023segloc}, Gaussian splatting feature fields~\cite{pietrantoni2025gsff}, and NeRF-based representations~\cite{pietrantoni2025ppnesf}, yet they rely on extensive priors or labels and limit generalizability.

\paragraph{Cross-Detector Matching.}
Standard localization pipelines assume the same feature extractor for mapping and querying. To enable cross-detector map reuse, prior works learn pairwise descriptor translations~\cite{dusmanu2021cross}, project features into shared embedding spaces~\cite{jaenal2025towards}, or augment descriptors with geometric context~\cite{cubero2025matcha}, all requiring descriptor storage or alignment. However, these methods must be retrained or fine-tuned for every new detector pair, limiting scalability as the number of detectors grows. In contrast, our descriptor-free formulation sidesteps the cross-detector problem entirely: since matching relies solely on geometric relations, keypoints from any detector are directly interchangeable without additional training or adaptation.

\begin{figure*}[!t]
    \centering
    \resizebox{0.95\textwidth}{!}{\input{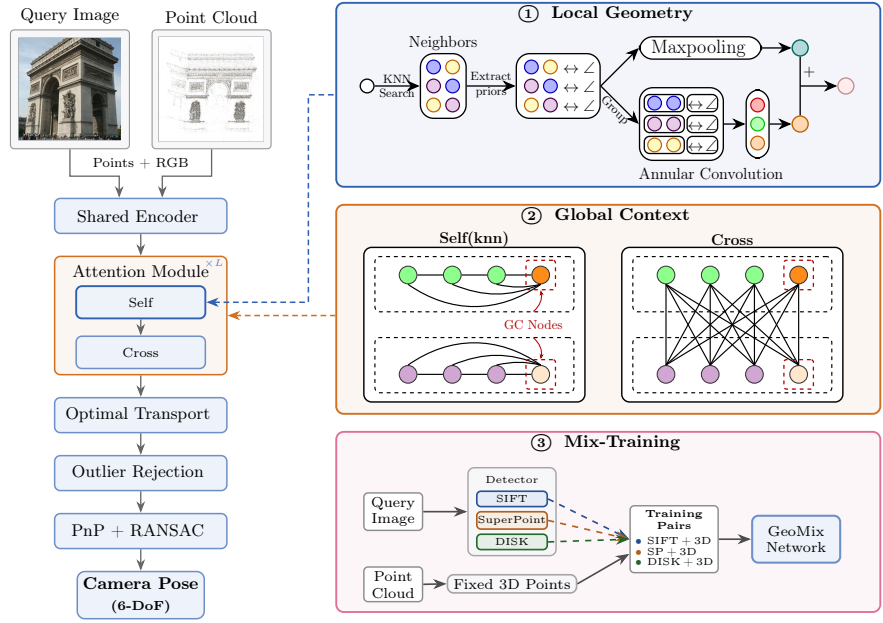}}%
    \caption{Overview of GeoMix. A shared encoder lifts 2D keypoints and 3D points to bearing vectors with RGB colors. The attention module alternates self-attention and cross-attention over $L$ layers: self-attention aggregates local geometry via annular convolution and max-pooling enriched by directional and distance embeddings, while learnable global context (GC) nodes broadcast scene-wide information through cross-attention. Optimal transport with outlier rejection produces 2D--3D correspondences for pose estimation. During training, Mix-Training varies the query-side keypoint detector while keeping the 3D map fixed, encouraging detector-agnostic geometric learning.}
    \label{fig:network}
\end{figure*}

\section{Method}
\label{sec:method}

\subsection{Problem Setting and Keypoint Representation}
\label{sec:problem_setting}

\paragraph{Problem Formulation.}
Given a query image $I$, we first retrieve its top-$k$ database images and collect the 3D points observed in them to form a candidate point cloud
$\mathbf{Q} = \{\mathbf{q}_j \in \mathbb{R}^3\}_{j=1}^{N}$.
Because the retrieved database images are registered in the scene, their known camera poses provide the extrinsic parameters used below.
We estimate the 6-DoF camera pose $(\mathbf{R}, \mathbf{t})$ of the query image by establishing 2D--3D correspondences between its $M$ keypoints
$\mathbf{P} = \{\mathbf{p}_i \in \mathbb{R}^2\}_{i=1}^{M}$
and $\mathbf{Q}$.

\paragraph{Keypoint Representation.}
Since 2D keypoints are defined in pixel space while 3D scene points reside in world coordinates, directly comparing them is challenging without a common representation. Following~\cite{zhou2022geometry,wang2024dgc,zhang2025a2gnnangleannulargnnvisual}, we adopt bearing vectors to bridge this modality gap: by back-projecting both onto the normalized image plane, 2D and 3D keypoints become geometrically comparable. For a 2D keypoint $\mathbf{p}_i = (u, v)$, we back-project it through the camera intrinsics $\mathbf{K}$ to obtain the bearing vector $\mathbf{b}_{p_i}$:
$$ [\mathbf{b}_{p_i}^\top, 1]^\top = \mathbf{K}^{-1} [u, v, 1]^\top. $$
For a 3D scene point $\mathbf{q}_j$ and the world-to-database-image transformation with rotation $\mathbf{R}$ and translation $\mathbf{t}$, we compute $\mathbf{q}_j^{c} = \mathbf{R} \mathbf{q}_j + \mathbf{t}$, where $\mathbf{q}_j^{c}$ represents $\mathbf{q}_j$ in the camera's frame of reference. We then obtain the bearing vector $\mathbf{b}_{q_j}$ by projecting onto the normalized image plane:
$$ [\mathbf{b}_{q_j}^\top, 1]^\top = \frac{\mathbf{q}_j^{c}}{q_{j,z}^{c}}, $$
where $q_{j,z}^{c}$ denotes the $z$-coordinate of $\mathbf{q}_j^{c}$.

\subsection{Network Architecture}
\label{sec:network_architecture}


\cref{fig:network} shows our architecture, which builds upon the descriptor-free A2-GNN backbone~\cite{zhang2025a2gnnangleannulargnnvisual}. The model consists of a bearing/RGB feature encoder, a geometric aggregation GNN with angle–annular convolution and max-pooling for local structure modeling, a global context mechanism, an optimal transport layer for differentiable correspondence estimation, and an outlier rejection module. While inheriting the encoder, local aggregation, optimal transport, and rejection modules from A2-GNN, our GeoMix introduces three key components in \cref{fig:teaser}(b-ii): directional and distance edge embeddings, learnable global context nodes, and a Mix-Training strategy. These components enhance local and global geometric embedding and improve robustness.

\paragraph{Feature Encoder.}
We use ResNet-style encoders to project inputs into a high-dimensional feature space. For each keypoint with bearing vector $\mathbf{b}_i \in \{\mathbf{b}_{p_i}, \mathbf{b}_{q_j}\}$ and RGB color $\mathbf{c}_i$, we process geometric and photometric cues through separate branches and sum their embeddings to generate the initial feature vector $\mathbf{f}_i \in \mathbb{R}^d$:
\begin{equation}
    \mathbf{f}_i = \mathcal{F}_b(\mathbf{b}_i) + \mathcal{F}_c(\mathbf{c}_i),
    \label{eq:feature_encoding}
\end{equation}
where $\mathcal{F}_b$ and $\mathcal{F}_c$ denote the bearing vector and color encoders.

\paragraph{Geometric Graph Neural Network.}
We employ a GNN where each node represents a 2D keypoint or 3D point. The network alternates between self-attention and cross-attention to progressively refine features. Self-attention layers extract local geometric context via dual-branch aggregation, while cross-attention layers incorporate global scene context using global context nodes.

\paragraph{Local Geometry.}
We construct a local graph by connecting each point to its $k$ nearest neighbors. For central node $i$ with neighbor set $\mathcal{N}_i$, self-attention uses dual-branch aggregation: geometry-guided max-pooling extracts salient metric cues with permutation invariance, while annular convolution partitions neighbors into distance-based groups to encode topological structure.
For each neighbor $j \in \mathcal{N}_i$, we compute relative displacement $\Delta\mathbf{p}_{ij} = \mathbf{p}_j - \mathbf{p}_i$ and form a unified geometric embedding shared by both branches by concatenating the relative displacement with its unit direction:
\begin{equation}
    \mathbf{e}_{ij}^{geo} = \left[ \Delta \mathbf{p}_{ij} \concat \frac{\Delta \mathbf{p}_{ij}}{\|\Delta \mathbf{p}_{ij}\|_2} \right] \in \mathbb{R}^4.
\end{equation}
This embedding is subsequently projected and aggregated via a max-pooling operation:
\begin{equation}
    \mathbf{h}_{i}^{max} = \max_{j \in \mathcal{N}_i} \left( \phi_{feat}([\mathbf{f}_i \concat \mathbf{f}_j - \mathbf{f}_i]) + \phi_{geo}(\mathbf{e}_{ij}^{geo}) \right),
\end{equation}
yielding a geometry-aware representation $\mathbf{h}_{i}^{max}$, where $\phi_{feat}$ and $\phi_{geo}$ denote MLPs with instance normalization and LeakyReLU.

Simultaneously, the annular convolution branch partitions the $k$ nearest neighbors into $g$ groups by ranking Euclidean distances to the central node, so that each group contains $k/g$ neighbors at a similar radius. Two successive 1D convolutions with kernels of size $1 \times (k/g)$ and $1 \times g$ are applied: the first aggregates features within each group, and the second aggregates across groups. We apply this operation to both the concatenated features and the geometric embedding in parallel:
\begin{equation}
      \mathbf{h}_{i}^{ann} = \phi_{ann}^{feat}([\mathbf{f}_i \concat \mathbf{f}_j - \mathbf{f}_i]) + \phi_{ann}^{geo}(\mathbf{e}_{ij}^{geo}),
\end{equation}
where $\phi_{ann}^{feat}$ and $\phi_{ann}^{geo}$ denote the annular convolutions applied to neighbor features and geometric embeddings, respectively, each followed by batch normalization and ReLU.

Finally, we aggregate local features from two complementary paths: first, an annular convolution path $\mathbf{h}_{i}^{ann}$ with annular geometric embeddings, and second, a maxpooling path $\mathbf{h}_{i}^{max}$ with geometric embeddings. These are fused via element-wise summation: $\mathbf{f}_{i}^{local} = \mathbf{h}_{i}^{ann} + \mathbf{h}_{i}^{max}$, which then feeds into the global context module.

\paragraph{Global Context Nodes.}
To capture long-range dependencies beyond the local KNN graph, we introduce $N_g$ learnable global context nodes $\{\mathbf{g}_k \in \mathbb{R}^d\}_{k=1}^{N_g}$, maintained independently for each 2D and 3D point set. These nodes aggregate information from all local features, exchange context among themselves, and broadcast back via cross-attention:

\begin{align}
\mathbf{g}'_k &= \mathbf{g}_k + \phi_{agg}\left( \left[ \mathbf{g}_k \concat \operatorname{CA}(\mathbf{g}_k, \mathcal{F}) \right] \right), \label{eq:agg} \\
\mathbf{g}''_k &= \mathbf{g}'_k + \operatorname{SA}(\mathbf{g}'_k, \mathcal{G}'), \label{eq:int} \\
\mathbf{f}_i^{self} &= \mathbf{f}_i^{local} + \phi_{broad}\left( \left[ \mathbf{f}_i^{local} \concat \operatorname{CA}(\mathbf{f}_i^{local}, \mathcal{G}'') \right] \right), \label{eq:broad}
\end{align}
where $\operatorname{SA}(\cdot)$ and $\operatorname{CA}(\cdot)$ denote Self-Attention and Cross-Attention mechanisms respectively. $\mathcal{G}'$ and $\mathcal{G}''$ denote the sets of intermediate global context features, $[\cdot \concat \cdot]$ denotes concatenation, and $\phi_{agg}$ and $\phi_{broad}$ are MLP blocks for aggregation and broadcasting, respectively. The final output $\mathbf{f}_i^{self}$ corresponds to the locally updated feature for node $i$ enriched with global context.


\paragraph{Optimal Transport and Outlier Rejection.}
To establish the initial set of correspondences $\mathcal{M}_{init}$, we employ a differentiable Optimal Transport layer. By applying the Sinkhorn algorithm~\cite{cuturi2013sinkhorn,sinkhorn1967concerning} to the score matrix derived from the enhanced features, we compute an optimal assignment matrix that encapsulates the soft matching probabilities between 2D query keypoints and 3D scene points.

Subsequently, following A2-GNN~\cite{zhang2025a2gnnangleannulargnnvisual}, we utilize a learning-based outlier rejection module to refine these initial matches. This classifier estimates a confidence score for each putative match, indicating its probability of being an inlier. The final set of robust correspondences is obtained by retaining only those matches where the predicted confidence exceeds a fixed threshold $\tau$.

\begin{figure*}[!t]
\centering
\begin{tikzpicture}
\def\w{0.245\textwidth}
\def\h{0.3\textheight}
\tikzset{every node/.style={inner sep=0pt, outer sep=0pt}}

\node (img2) [anchor=west] {
    \includegraphics[width=\w,height=\h,keepaspectratio]{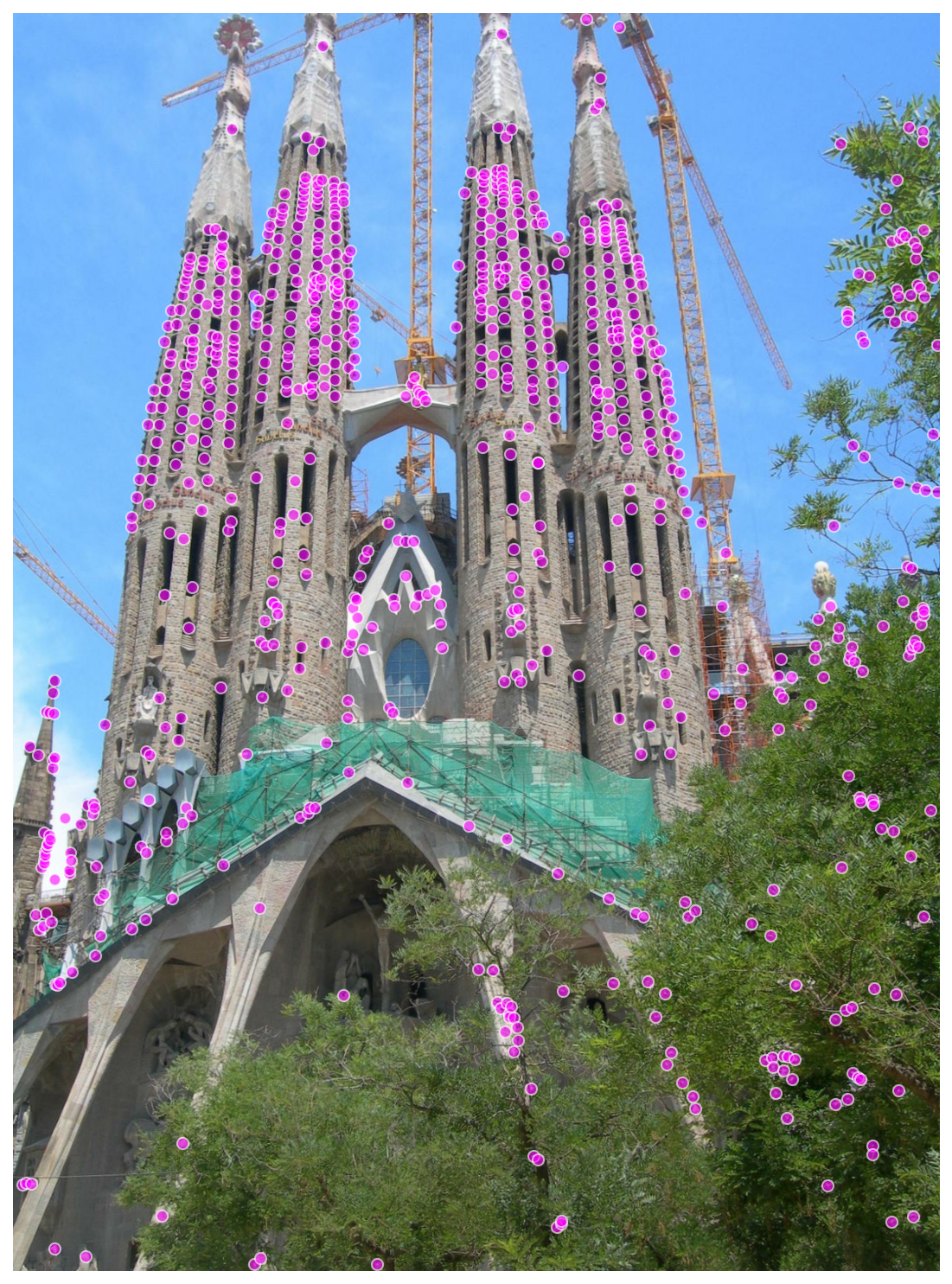}
};
\node (img3) [anchor=west, right=0pt of img2.east] {
    \includegraphics[width=\w,height=\h,keepaspectratio]{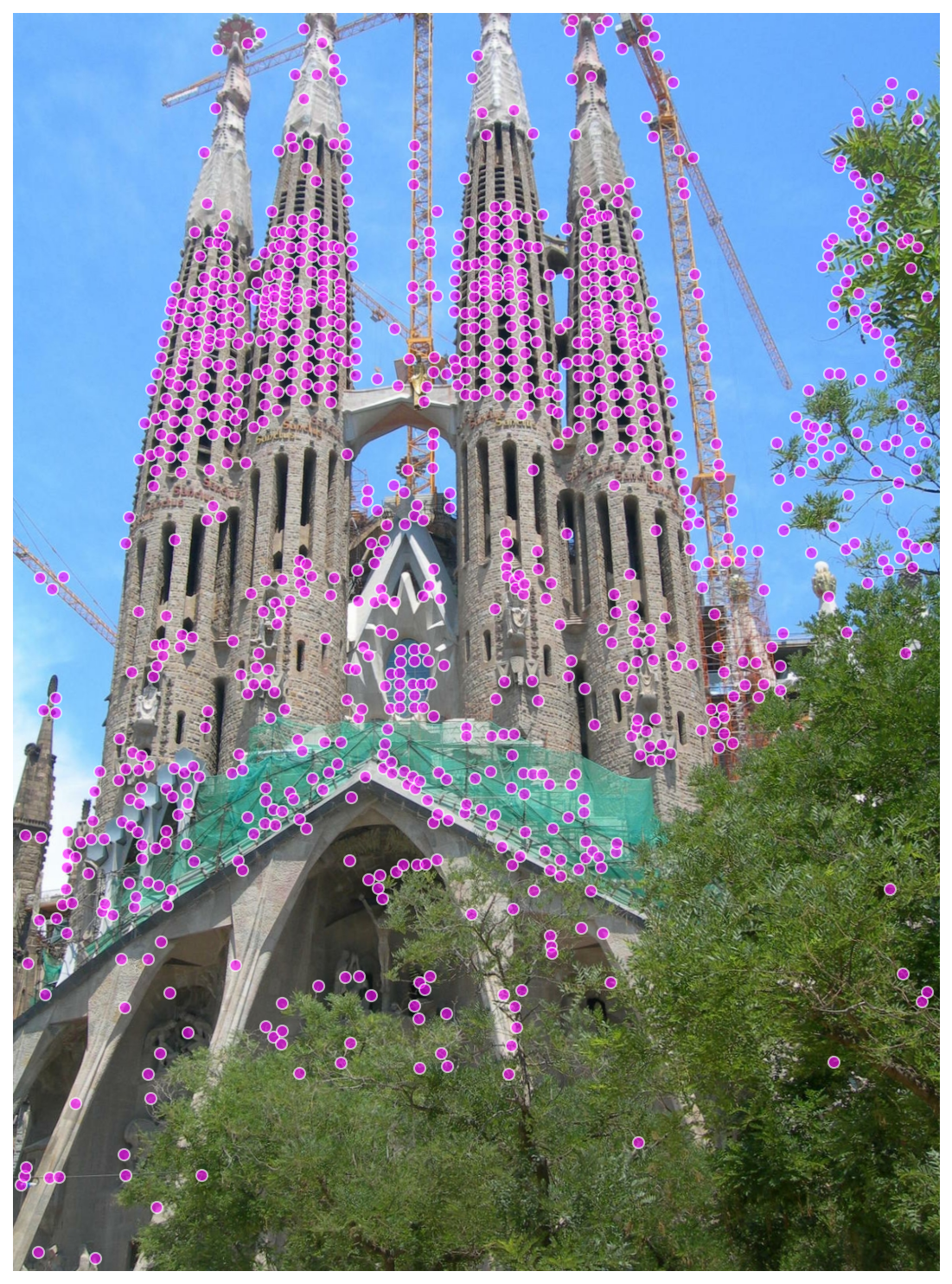}
};
\node (img4) [anchor=west, right=0pt of img3.east] {
    \includegraphics[width=\w,height=\h,keepaspectratio]{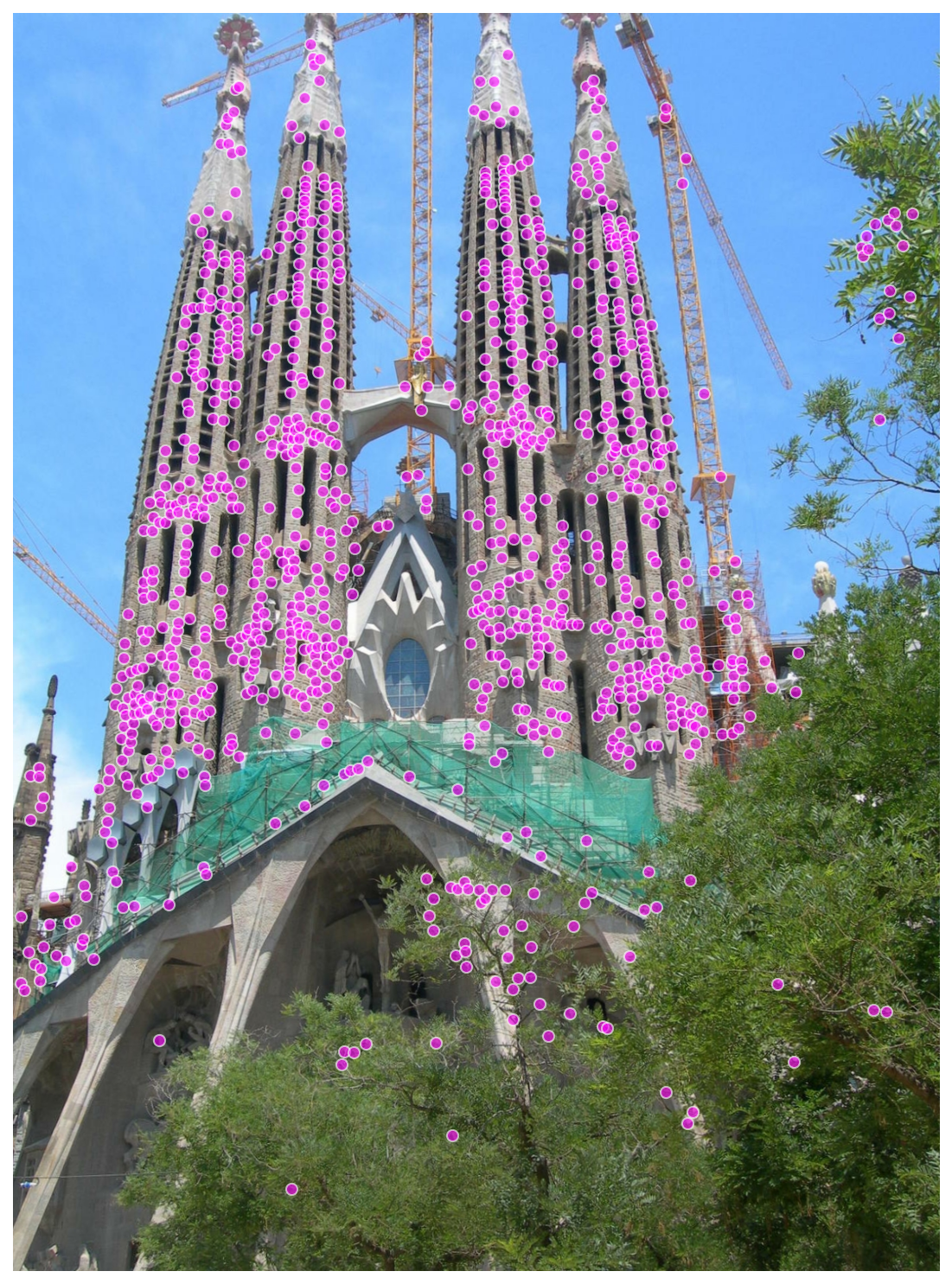}
};
\node (img5) [anchor=west, right=0pt of img4.east] {
    \includegraphics[width=\w,height=\h,keepaspectratio]{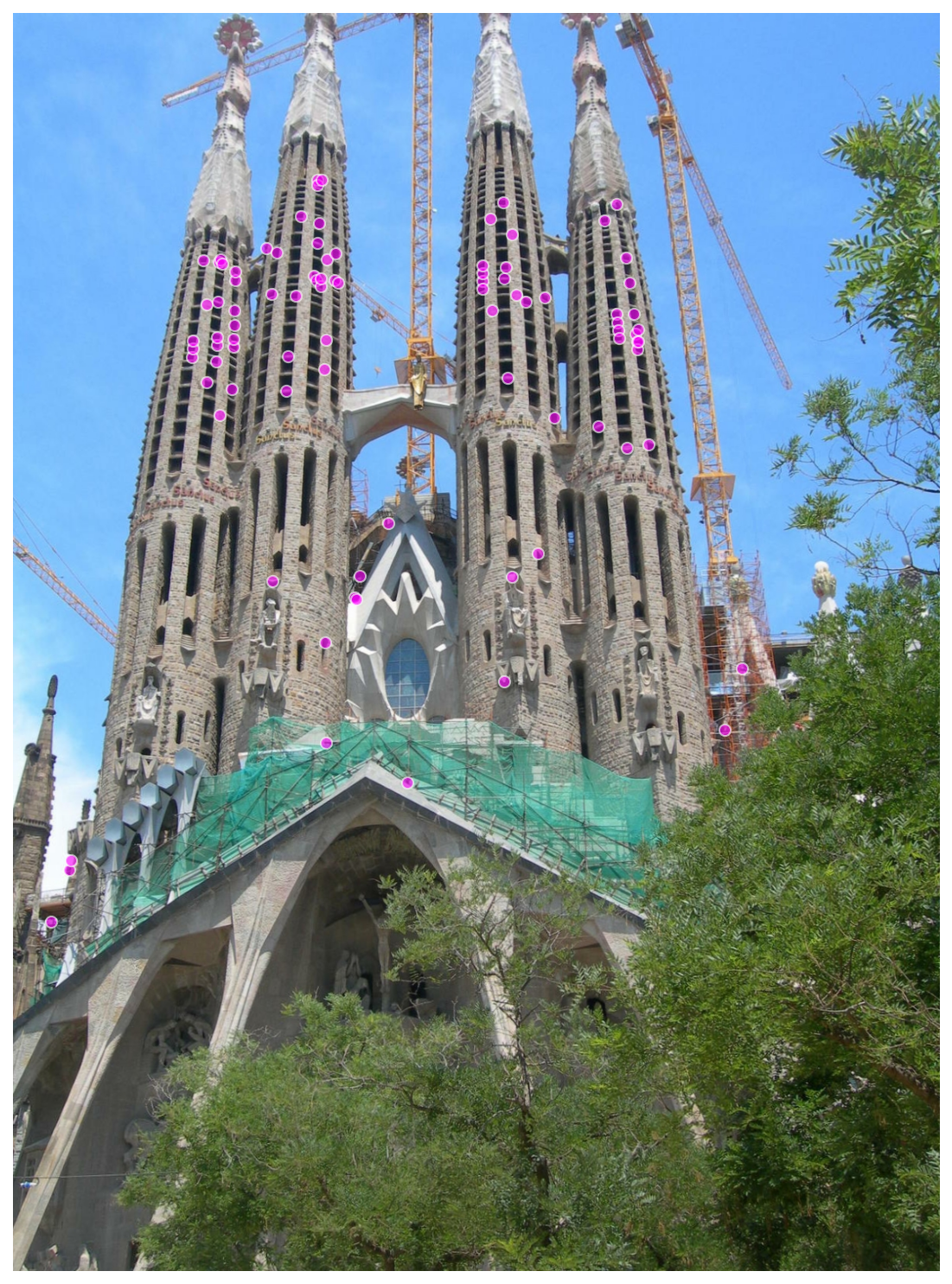}
};

\node[below=2mm of img2.south] {\footnotesize SIFT};
\node[below=2mm of img3.south] {\footnotesize SuperPoint};
\node[below=2mm of img4.south] {\footnotesize DISK};
\node[below=2mm of img5.south] {\footnotesize Shared Points};
\end{tikzpicture}
\caption{Keypoint distributions of SIFT, SuperPoint, and DISK on the same image. Each detector produces a distinct spatial layout with few shared keypoints (rightmost), motivating Mix-Training to learn detector-agnostic geometric representations.}
\label{fig:multi}
\end{figure*}

\subsection{Mix-Training Strategy}

Different keypoint detectors adopt different detection strategies and produce markedly different keypoint distributions. As shown in \cref{fig:multi}, SIFT~\cite{lowe2004distinctive} favors blob-like structures, SuperPoint~\cite{detone2018superpoint} prefers corner regions, and DISK~\cite{tyszkiewicz2020disk} targets repeatable local features, resulting in few shared keypoints on the same image. However, existing descriptor-free methods are typically trained with a single detector, causing the network to overfit to that detector's keypoint distribution and generalize poorly to others.

To address this, we introduce Mix-Training, a multi-detector training strategy uniquely enabled by the descriptor-free setting. In descriptor-based methods, each detector is tightly coupled with its own descriptor space, making it impractical to mix detectors without reconciling incompatible representations. The descriptor-free formulation removes this constraint: since matching is based purely on geometric relationships between 2D-3D correspondences, keypoints from any detector can serve as input without modifying the matching pipeline.

Specifically, we maintain a pool of $L$ detectors $\mathcal{S} = \{D_l\}_{l=1}^{L}$ ($L{=}3$ in practice). For each query image, we apply all $L$ detectors independently to generate $L$ distinct sets of 2D keypoints. Each set is paired with the same 3D scene points to produce $L$ training samples per image. Note that we fix the 3D side and vary only the 2D query-side detector, since the 3D maps are reconstructed via SfM with SIFT and regenerating them for each detector would be prohibitively expensive. By exposing the network to diverse keypoint distributions against shared 3D geometry, Mix-Training encourages detector-invariant geometric reasoning and improves generalization to unseen detectors at test time. At inference a single detector still yields 1,024 keypoints per query, identical to single-detector training, so the gains stem from detector diversity rather than a larger keypoint budget.

\subsection{Training Loss}
Following GoMatch~\cite{zhou2022geometry}, the loss function $\mathcal{L}$ consists of a matching loss $\mathcal{L}_{m}$ and an outlier rejection loss $\mathcal{L}_{or}$. The matching loss minimizes the negative log-likelihood of matching scores:
\begin{align}
    \mathcal{L}_{m} = - \frac{1}{N_m} (\sum_{(i, j) \in \mathcal{M}_{gt}} \log \tilde{\text{P}}_{ij} +  \sum_{i \in \mathcal{U}_{q}} \log \tilde{\text{P}}_{i(N + 1)} + \sum_{j \in \mathcal{U}_{d}} \log \tilde{\text{P}}_{(M + 1)j} ), 
\end{align}
where $\tilde{\text{P}}$ denotes the matching score matrix, $\mathcal{M}_{gt}$ the set of ground-truth matches, $\mathcal{U}_{q}$ the unmatched query keypoints, and $\mathcal{U}_{d}$ the unmatched database 3D points. $N_m$ is the total number of ground-truth, unmatched query, and unmatched database keypoints.

The outlier rejection loss $\mathcal{L}_{or}$ enhances robustness against outliers:
\begin{equation}
    \mathcal{L}_{or} = - \frac{1}{N_c}\sum_{i=1}^{N_c} w_i \left(y_i \log p_i + (1 - y_i) \log (1 - p_i) \right),
\end{equation}
where $N_c$ denotes the total number of initial correspondences, $p_i$ is the predicted inlier probability, $y_i$ the ground-truth label, and $w_i$ the class-balancing weight. The total training loss is $\mathcal{L} = \mathcal{L}_{m} + \mathcal{L}_{or}$.

\begin{table*}[!t]
\begin{center}
\caption{Matching results on MegaDepth under two training protocols. \textit{Single-training} uses SIFT only; \textit{Mix-training} jointly trains with SIFT, SuperPoint, and DISK. The upper section evaluates detectors seen during training, while the lower section tests zero-shot transfer to unseen detectors (DeDoDe v2 and R2D2). Best results are in \textbf{bold}.}
\label{tab: matching_detectors}
\renewcommand\arraystretch{1.2}
\setlength{\tabcolsep}{4pt}
\scriptsize
\begin{tabular}{llccc}
\toprule
\multirow{2}{*}{Method} &
\multirow{2}{*}{Detector} &
  Reproj. AUC (\%) &
  Rotation ($^\circ$)  &
  Translation \\

 & &
  @1 / 5 / 10px ($\uparrow$) &
  \multicolumn{2}{c}{Quantile @25 / 50 / 75\% $(\downarrow)$}
   \\ \midrule

\multirow{3}{*}{Single-training} & SIFT       & 18.68 / 58.34 / 66.19 & 0.06 / 0.15 / \phantom{1}2.26 & 0.00 / 0.01 / 0.20 \\
& SuperPoint & \phantom{1}0.04 / \phantom{1}8.55 / 17.50 & 1.67 / 7.14 / 27.40 & 0.15 / 0.68 / 2.94 \\
& DISK       & \phantom{1}0.02 / 17.70 / 32.00 & 0.74 / 2.82 / 16.06 & 0.06 / 0.25 / 1.69 \\
   \midrule

\multirow{3}{*}{Mix-training} & SIFT       & \B{21.51} / \B{65.61} / \B{73.76} & \B{0.05} / \B{0.11} / \phantom{1}\B{0.52} & \B{0.00} / \B{0.01} / \B{0.05} \\
& SuperPoint & \phantom{1}\B{0.15} / \B{20.84} / \B{36.89} & \B{0.60} / \B{1.57} / \phantom{1}\B{7.57} & \B{0.05} / \B{0.14} / \B{0.77} \\
& DISK       & \phantom{1}\B{0.04} / \B{30.30} / \B{48.80} & \B{0.33} / \B{0.96} / \phantom{1}\B{4.22} & \B{0.03} / \B{0.08} / \B{0.41} \\ \midrule
\multirow{2}{*}{Single-training} & DeDoDe v2 & \phantom{1}7.87 / 47.27 / 57.31 & 0.11 / 0.39 / \phantom{1}4.19 & 0.01 / 0.03 / 0.44 \\
& R2D2      & \phantom{1}0.29 / 27.29 / 37.62 & 0.29 / 1.94 / 17.07 & 0.03 / 0.18 / 1.89 \\ \midrule
\multirow{2}{*}{Mix-training} & DeDoDe v2 & \B{11.16} / \B{60.30} / \B{70.56} & \B{0.07} / \B{0.17} / \phantom{1}\B{0.83} & \B{0.01} / \B{0.02} / \B{0.07} \\
& R2D2      & \phantom{1}\B{0.62} / \B{46.87} / \B{60.22} & \B{0.13} / \B{0.34} / \phantom{1}\B{1.95} & \B{0.01} / \B{0.03} / \B{0.19} \\ \bottomrule
\end{tabular}
\end{center}
\end{table*}

\section{Experiments}
\label{sec:experiments}

\paragraph{Training.}
We train our model on the MegaDepth dataset~\cite{li2018megadepth} for 50 epochs using the Adam optimizer~\cite{kingma2014adam} with a learning rate of $10^{-3}$ and a batch size of 16. We use $N_g{=}4$ global context nodes, $9$ nearest neighbors for local graph construction, and a group size of 3 for neighborhood aggregation, following the same settings as A2-GNN~\cite{zhang2025a2gnnangleannulargnnvisual}. Training runs on a single NVIDIA RTX 4090 GPU with 24\,GB of VRAM and converges in approximately 22 hours.

\begin{table}[!t]
\begin{center}
\caption{Visual localization on Cambridge Landmarks. Median translation (cm) and rotation ($^\circ$) errors are reported. MB denotes total map size. Best results per group are in \textbf{bold}.}
\label{tab:cambridge}
\renewcommand\arraystretch{1.2}
\setlength{\tabcolsep}{4pt}
\scriptsize
\begin{tabular}{llcccccc}
\toprule
\multicolumn{2}{l}{\multirow{2}{*}{\textbf{Methods}}} &
  \multicolumn{6}{c}{\textbf{Cambridge-Landmarks} (cm, $^\circ$)} \\
\cmidrule(lr){3-8}
\multicolumn{2}{l}{} &
  King's & Hospital & Shop & St. Mary & Avg. & \textbf{MB} \\
\midrule
 \multirow{4}{*}{\rotatebox{90}{\textbf{E2E}}} &
  MS-Trans.~\cite{shavit2021learning} &
  83 / 1.47 & 181 / 2.39 & 86 / 3.07 & 162 / 3.99 & 128 / 2.73 & \phantom{00}\B{71} \\
 &
  DSAC*~\cite{brachmann2021visual} &
  \B{15} / \B{0.30} & \phantom{0}21 / 0.40 & \B{\phantom{0}5} / \B{0.30} & \phantom{0}13 / 0.40 & \phantom{0}14 / 0.35 & \phantom{0}112 \\
 &
  HSCNet~\cite{li2020hierarchical} &
  18 / \B{0.30} & \phantom{0}\B{19} / \B{0.30} & \phantom{0}6 / \B{0.30} & \phantom{00}\B{9} / \B{0.30} & \B{13} / \B{0.30} & \phantom{0}592 \\
 &
  Reloc3r~\cite{dong2025reloc3r} &
  42 / 0.36 & \phantom{0}62 / 0.55 & 13 / 0.58 & \phantom{0}34 / 0.58 & \phantom{0}38 / 0.52 & 10294 \\
\midrule
 \multirow{3}{*}{\rotatebox{90}{\textbf{DB}}} &
 HybridSC~\cite{camposeco2019hybrid} &
  81 / 0.59 & \phantom{0}75 / 1.01 & 19 / 0.54 & \phantom{0}50 / 0.49 & \phantom{0}56 / 0.66 & \phantom{000}\B{3} \\
 &
  AS~\cite{sattler2016efficient} &
  13 / 0.22 & \phantom{0}20 / 0.36 & \B{\phantom{0}4} / \B{0.21} & \phantom{00}8 / 0.25 & \phantom{0}11 / 0.26 & \phantom{0}813 \\
 &
  SP~\cite{detone2018superpoint}+SG~\cite{sarlin2020superglue} &
  \B{12} / \B{0.20} & \phantom{0}\B{15} / \B{0.30} & \B{\phantom{0}4} / \B{0.20} & \phantom{00}\B{7} / \B{0.21} & \B{10} / \B{0.23} & 3215 \\
\midrule
 \multirow{4}{*}{\rotatebox{90}{\textbf{DF}}}
 &
  GoMatch~\cite{zhou2022geometry} &
  25 / 0.64 & 283 / 8.14 & 48 / 4.77 & 335 / 9.94 & 173 / 5.87 & \phantom{00}\B{48} \\
 &
  DGC-GNN~\cite{wang2024dgc} &
  18 / 0.47 & \phantom{0}75 / 2.83 & 15 / 1.57 & 106 / 4.03 & \phantom{0}54 / 2.23 & \phantom{00}69 \\
 &
 A2-GNN~\cite{zhang2025a2gnnangleannulargnnvisual} &
  15 / 0.39 & \phantom{0}59 / 1.74 & 12 / 1.16 & \phantom{0}76 / 2.65 & \phantom{0}41 / 1.49 & \phantom{00}69 \\
 &
 GeoMix (Ours) &
  \B{14} / \B{0.37} & \phantom{0}\B{27} / \B{0.82} & \B{\phantom{0}6} / \B{0.73} & \phantom{0}\B{23} / \B{0.72} & \phantom{0}\B{18} / \B{0.66} & \phantom{00}69 \\
\bottomrule
\end{tabular}
\end{center}
\end{table}

\paragraph{Datasets.}
Following prior work~\cite{zhou2022geometry}, we train and evaluate on MegaDepth~\cite{li2018megadepth}, a large-scale outdoor dataset with SfM-based 3D models. We follow GoMatch~\cite{zhou2022geometry} and use the official test split of 53 sequences, with the remaining sequences divided into 99 training, 16 validation, and 49 test scenes after query sampling and co-visibility filtering. During training, we extract 2D keypoints with SIFT, SuperPoint, and DISK, and establish 2D-3D correspondences against the original SIFT-based SfM point clouds. For visual localization, we evaluate on Cambridge Landmarks~\cite{kendall2015posenet}, a medium-scale outdoor dataset with four scenes and SfM-based 3D reconstructions; 7Scenes~\cite{shotton2013scene}, a small-scale indoor dataset with seven RGB-D scenes and SLAM-derived ground-truth poses; and Aachen Day-Night~\cite{sattler2018benchmarking}, which requires localizing 922 queries against a 3D model built from 4,328 daytime database images under drastic day-night illumination changes. Detailed descriptions of the correspondence retrieval strategy and the Aachen Day-Night retriangulation procedure are provided in the supplementary material.

\paragraph{Keypoint Detectors.}
Unless otherwise specified, our main model uses mix-training on MegaDepth with three 2D query-side keypoint detectors: SIFT~\cite{lowe2004distinctive}, SuperPoint~\cite{detone2018superpoint}, and DISK~\cite{tyszkiewicz2020disk}, covering both classical hand-crafted and modern learning-based approaches. Note that the 3D point clouds in MegaDepth are reconstructed with SIFT, so training with SuperPoint or DISK introduces a cross-detector setting between the 2D and 3D sides. To isolate the effect of detector diversity, \cref{tab: matching_detectors} also includes a single-training baseline trained with SIFT only. For evaluation on MegaDepth, we test with SIFT, SuperPoint, and DISK, and further evaluate on R2D2~\cite{revaud2019r2d2} and DeDoDe-v2~\cite{edstedt2024dedodev2} as zero-shot detectors to test generalizability to unseen feature extractors. For visual localization, Cambridge Landmarks uses SuperPoint and 7Scenes uses SIFT, while Aachen Day-Night is evaluated with retriangulated 3D models from multiple detectors, following the configurations described above.

\paragraph{Evaluation Metrics.}
Following~\cite{zhou2022geometry, wang2024dgc,zhang2025a2gnnangleannulargnnvisual}, we adopt different metrics for matching and visual localization tasks. For matching evaluation on MegaDepth, we report the Area Under the Curve (AUC) of mean reprojection error at 1, 5, and 10 pixel thresholds, along with translation and rotation error quantiles at 25\%, 50\%, and 75\%. For visual localization on Cambridge Landmarks and 7Scenes, we report the median translation (cm) and rotation ($^\circ$) errors per scene. All camera poses are estimated using PnP-RANSAC~\cite{fischler1981random, gao2003complete} on the established correspondences. We also provide Oracle results, which use ground truth matches as an upper bound for reference.

\begin{table*}[!t]
\begin{center}
\caption{Visual localization on 7Scenes. Median translation (cm) and rotation ($^\circ$) errors are reported. MB denotes total map size. Best results per group are in \textbf{bold}.}
\label{tab:7scenes}
\renewcommand\arraystretch{1.2}
\setlength{\tabcolsep}{3pt}
\scriptsize
\resizebox{\textwidth}{!}{
\begin{tabular}{llccccccccc}
\toprule
\multicolumn{2}{l}{\multirow{2}{*}{\textbf{Methods}}} &
  \multicolumn{9}{c}{\textbf{7Scenes} (cm, $^\circ$) (SIFT)} \\
\cmidrule(lr){3-11}
\multicolumn{2}{l}{} &
  Chess & Fire & Heads & Office & Pumpkin & Kitchen & Stairs & Avg. & \textbf{MB} \\
\midrule
 \multirow{4}{*}{\rotatebox{90}{\textbf{E2E}}} &
  MS-Trans.~\cite{shavit2021learning} & 11/4.66 & 24/9.60 & 14/12.2\phantom{0} & 17/5.66 & 18/4.44 & 17/5.94 & 26/\phantom{0}8.45 & 18/7.28 & \phantom{000}\B{71} \\
 & DSAC*~\cite{brachmann2021visual} & \phantom{0}\B{2}/1.10 & \phantom{0}\B{2}/1.24 & \phantom{0}\B{1}/\phantom{0}1.82 & \phantom{0}\B{3}/1.15 & \phantom{0}\B{4}/1.34 & \phantom{0}\B{4}/1.68 & \phantom{0}\B{3}/\phantom{0}1.16 & \phantom{0}\B{3}/1.36 & \phantom{00}196 \\
 & HSCNet~\cite{li2020hierarchical} & \phantom{0}\B{2}/\B{0.70} & \phantom{0}\B{2}/\B{0.90} & \phantom{0}\B{1}/\phantom{0}\B{0.90} & \phantom{0}\B{3}/\B{0.80} & \phantom{0}\B{4}/\B{1.00} & \phantom{0}\B{4}/\B{1.20} & \phantom{0}\B{3}/\phantom{0}\B{0.80} & \phantom{0}\B{3}/\B{0.90} & \phantom{0}1036 \\
 & Reloc3r~\cite{dong2025reloc3r} & \phantom{0}3/0.88 & \phantom{0}3/0.81 & \phantom{0}\B{1}/\phantom{0}0.95 & \phantom{0}4/0.88 & \phantom{0}6/1.10 & \phantom{0}4/1.26 & \phantom{0}7/\phantom{0}1.26 & \phantom{0}4/1.02 & 10268 \\
\midrule
 \multirow{2}{*}{\rotatebox{90}{\textbf{DB}}} &
  AS~\cite{sattler2016efficient} & \phantom{0}3/0.87 & \phantom{0}\B{2}/1.01 & \phantom{0}\B{1}/\phantom{0}0.82 & \phantom{0}4/1.15 & \phantom{0}7/1.69 & \phantom{0}5/1.72 & \phantom{0}\B{4}/\phantom{0}\B{1.01} & \phantom{0}4/1.18 & - \\
 & SP~\cite{detone2018superpoint}+SG~\cite{sarlin2020superglue} & \phantom{0}\B{2}/\B{0.85} & \phantom{0}\B{2}/\B{0.94} & \phantom{0}\B{1}/\phantom{0}\B{0.75} & \phantom{0}\B{3}/\B{0.92} & \phantom{0}\B{5}/\B{1.30} & \phantom{0}\B{4}/\B{1.40} & \phantom{0}5/\phantom{0}1.47 & \phantom{0}\B{3}/\B{1.09} & 22977 \\
\midrule
 \multirow{4}{*}{\rotatebox{90}{\textbf{DF}}} &
  GoMatch~\cite{zhou2022geometry} & \phantom{0}4/1.65 & 13/3.86 & \phantom{0}9/\phantom{0}5.17 & 11/2.48 & 16/3.32 & 13/2.84 & 89/21.1\phantom{0} & 22/5.77 & \phantom{00}\B{302} \\
 & DGC-GNN~\cite{wang2024dgc} & \phantom{0}\B{3}/1.41 & \phantom{0}5/1.81 & \phantom{0}4/\phantom{0}3.13 & \phantom{0}7/1.66 & \phantom{0}8/2.03 & \phantom{0}8/2.14 & 83/21.5\phantom{0} & 17/4.81 & \phantom{00}355 \\
 & A2-GNN~\cite{zhang2025a2gnnangleannulargnnvisual} & \phantom{0}\B{3}/1.37 & \phantom{0}5/1.78 & \phantom{0}4/\phantom{0}2.70 & \phantom{0}6/1.56 & \phantom{0}\B{7}/1.86 & \phantom{0}\B{7}/2.00 & 72/17.0\phantom{0} & 15/4.04 & \phantom{00}355 \\
 & GeoMix (Ours) & \phantom{0}\B{3}/\B{1.31} & \phantom{0}\B{4}/\B{1.52} & \phantom{0}\B{3}/\phantom{0}\B{2.30} & \phantom{0}\B{5}/\B{1.46} & \phantom{0}\B{7}/\B{1.74} & \phantom{0}\B{7}/\B{1.88} & \B{45}/\B{11.1}\phantom{0} & \B{11}/\B{3.04} & \phantom{00}{355} \\
\bottomrule
\end{tabular}}
\end{center}
\end{table*}

\begin{table}[!t]
\begin{center}
\caption{Visual localization on Aachen Day-Night~\cite{sattler2018benchmarking}. Percentage of correctly localized queries at (0.25m, 2$^\circ$) / (0.5m, 5$^\circ$) / (5m, 10$^\circ$) thresholds. For descriptor-free methods, the 2D and 3D sides use the same detector type; SP+SG is a descriptor-based reference evaluated on the same retriangulated map. Matching time per image excludes PnP-RANSAC; Params is model size in millions. Best results among descriptor-free methods are in \textbf{bold}.}
\label{tab:aachen}
\renewcommand\arraystretch{1.2}
\setlength{\tabcolsep}{4pt}
\scriptsize
\begin{tabular}{llcccc}
\toprule
\textbf{Method} & \textbf{Detector} & \textbf{Day} & \textbf{Night} & \textbf{Time (s)} & \textbf{Params (M)} \\
\midrule
\multicolumn{6}{l}{\textit{Descriptor-based reference}} \\
SP~\cite{detone2018superpoint}+SG~\cite{sarlin2020superglue} & SuperPoint & 85.1 / 91.6 / 96.0 & 77.6 / 86.7 / 94.9 & 0.138 & 12.04 \\
\midrule
\multicolumn{6}{l}{\textit{Descriptor-free}} \\
\multirow{4}{*}{GoMatch~\cite{zhou2022geometry}}
 & SIFT  & \phantom{0}0.5 / \phantom{0}1.2 / \phantom{0}6.9 & 0.0 / \phantom{0}0.0 / \phantom{0}1.0 & \multirow{4}{*}{\B{0.077}} & \multirow{4}{*}{\B{1.32}} \\
 & SuperPoint    & \phantom{0}2.2 / \phantom{0}4.7 / 18.4 & 1.0 / \phantom{0}2.0 / 10.2 & & \\
 & R2D2  & \phantom{0}2.3 / \phantom{0}4.2 / 10.2 & 1.0 / \phantom{0}3.1 / \phantom{0}3.1 & & \\
 & DISK  & \phantom{0}4.2 / \phantom{0}7.9 / 26.3 & 0.0 / \phantom{0}1.0 / \phantom{0}4.1 & & \\
\midrule
\multirow{4}{*}{DGC-GNN~\cite{wang2024dgc}}
 & SIFT  & \phantom{0}2.8 / \phantom{0}6.3 / 15.9 & 0.0 / \phantom{0}0.0 / \phantom{0}1.0 & \multirow{4}{*}{0.284} & \multirow{4}{*}{5.65} \\
 & SuperPoint    & \phantom{0}8.1 / 13.3 / 31.1 & 0.0 / \phantom{0}2.0 / \phantom{0}7.1 & & \\
 & R2D2  & \phantom{0}4.6 / \phantom{0}8.1 / 16.0 & 1.0 / \phantom{0}1.0 / \phantom{0}3.1 & & \\
 & DISK  & 11.7 / 20.6 / 38.5 & 1.0 / \phantom{0}2.0 / \phantom{0}9.2 & & \\
\midrule
\multirow{4}{*}{A2-GNN~\cite{zhang2025a2gnnangleannulargnnvisual}}
 & SIFT  & \phantom{0}3.8 / \phantom{0}7.5 / 16.7 & 0.0 / \phantom{0}0.0 / \phantom{0}3.1 & \multirow{4}{*}{0.108} & \multirow{4}{*}{2.67} \\
 & SuperPoint    & 11.2 / 20.3 / 36.9 & 1.0 / \phantom{0}3.1 / 11.2 & & \\
 & R2D2  & \phantom{0}6.4 / 11.4 / 19.5 & 1.0 / \phantom{0}3.1 / \phantom{0}4.1 & & \\
 & DISK  & 16.1 / 24.6 / 41.7 & 0.0 / \phantom{0}1.0 / \phantom{0}7.1 & & \\
\midrule
\multirow{4}{*}{GeoMix (Ours)}
 & SIFT  & \phantom{0}\B{7.9} / \B{15.9} / \B{32.6} & \B{2.0} / \phantom{0}\B{4.1} / \phantom{0}\B{5.1} & \multirow{4}{*}{0.115} & \multirow{4}{*}{3.50} \\
 & SuperPoint    & \B{25.6} / \B{37.4} / \B{58.1} & \B{9.2} / \B{16.3} / \B{27.6} & & \\
 & R2D2  & \B{11.0} / \B{18.4} / \B{33.1} & \B{3.1} / \phantom{0}\B{4.1} / \phantom{0}\B{7.1} & & \\
 & DISK  & \B{27.8} / \B{42.5} / \B{61.4} & \B{4.1} / \phantom{0}\B{8.2} / \B{12.2} & & \\
\bottomrule
\end{tabular}
\end{center}
\end{table}

\paragraph{Matching and Localization Results.}
\cref{tab: matching_detectors} compares single-training (SIFT only) with mix-training (SIFT, SuperPoint, DISK) under different test-time detectors. Single-training suffers severe cross-detector degradation, revealing a strong detector-specific bias. Mix-training resolves this: it not only narrows the cross-detector gap dramatically, but also improves same-detector performance, suggesting that diverse detector exposure acts as a regularizer that strengthens geometric reasoning.

We further compare against prior descriptor-free methods on MegaDepth with $k{=}10$ retrieved images in \cref{tab:mega_k10}. GeoMix outperforms all baselines by a substantial margin, improving AUC@5/10px over A2-GNN by +11.20/+11.52\%. The pose estimation gains are equally striking: the 75\%-quantile rotation error drops from $4.60^\circ$ to $0.52^\circ$ and the translation error from 0.48 to 0.05, both roughly 90\% reductions. To assess matcher quality independently of PnP-RANSAC, we further report matcher-only metrics (precision, recall, F1) on MegaDepth in the supplementary, where GeoMix attains the best scores on all three, consistent with the pose-based ranking.

To verify cross-dataset transfer, we evaluate on Cambridge Landmarks and 7Scenes as shown in \cref{tab:cambridge,tab:7scenes}. GeoMix achieves state-of-the-art results among descriptor-free methods on all scenes, with the largest gains on challenging cases such as Old Hospital, where translation error drops to 27\,cm, 54\% lower than A2-GNN, and Stairs, where GeoMix achieves 45\,cm\,/\,11.1$^\circ$ vs.\ 72\,cm\,/\,17.0$^\circ$ for A2-GNN. These results substantially narrow the gap with descriptor-based methods such as SuperPoint+SuperGlue, while requiring only 69\,MB storage versus 3215\,MB and preserving privacy by design. Scene-coordinate regressors such as DSAC* are also compact, but must be retrained per scene (hours each), whereas GeoMix is trained once on MegaDepth and reused zero-shot; even 32-d compact descriptors would still need roughly $6\times$ more storage than our geometry-only map.

We further stress-test robustness under drastic appearance changes on Aachen Day-Night in \cref{tab:aachen}, where both the 2D and 3D sides use the same detector type via retriangulation. GeoMix consistently outperforms all baselines across all four detectors, with especially large gains under the challenging nighttime setting, demonstrating strong robustness to illumination variation.

\begin{table}[!t]
\begin{center}
\caption{Comparison on MegaDepth ($k{=}10$ retrieved images). Best results are in \textbf{bold}.}
\label{tab:mega_k10}
\renewcommand\arraystretch{1.2}
\setlength{\tabcolsep}{4pt}
\scriptsize
\begin{tabular}{lccc}
\toprule
\multirow{2}{*}{Method} &
  Reproj. AUC (\%) &
  Rotation ($^\circ$) &
  Translation \\
& @1 / 5 / 10px ($\uparrow$) &
  \multicolumn{2}{c}{Quantile @25 / 50 / 75\% ($\downarrow$)} \\
\midrule
Oracle & 34.59 / 85.02 / 92.02 & 0.04 / 0.06 / \phantom{0}0.12 & 0.00 / 0.01 / 0.01 \\
\midrule
GoMatch~\cite{zhou2022geometry} & \phantom{0} 8.90 / 35.67 / 44.99 & 0.18 / 1.29 / 16.65 & 0.02 / 0.12 / 1.92 \\
DGC-GNN~\cite{wang2024dgc} & 15.30 / 51.70 / 60.01 & 0.07 / 0.26 / \phantom{0}5.41 & 0.01 / 0.02 / 0.57 \\
A2-GNN~\cite{zhang2025a2gnnangleannulargnnvisual} & 17.29 / 54.41 / 62.24 & 0.06 / 0.19 / \phantom{0}4.60 & 0.01 / 0.02 / 0.48 \\
GeoMix (Ours) & \B{21.51 / 65.61 / 73.76} & \B{0.05 / 0.11 / \phantom{0}0.52} & \B{0.00 / 0.01 / 0.05} \\
\bottomrule
\end{tabular}
\end{center}
\end{table}

\begin{table*}[!t]
\centering
\caption{Ablation study on MegaDepth evaluating the contribution of each proposed component: GC nodes, local geometry embeddings, and Mix-Training. Best results are in \textbf{bold}.}
\label{tab:abla}
\renewcommand\arraystretch{1.2}
\small
\resizebox{0.98\textwidth}{!}{
\begin{tabular}{lcccc|c|c|c}
\toprule
\multirow{2}{*}{Methods} &
\multirow{2}{*}{GC Node} &
\multicolumn{2}{c}{Local Geometry} &
\multirow{2}{*}{Mix Train} &
\multicolumn{1}{c}{Reproj. AUC (\%)} &
\multicolumn{1}{c}{Rotation ($^\circ$)} &
\multicolumn{1}{c}{Translation} \\
\cmidrule(lr){3-4}
& & Annular Conv. & Maxpooling & &
@1 / 5 / 10px ($\uparrow$) &
\multicolumn{2}{c}{Quantile@25 / 50 / 75\% ($\downarrow$)} \\
\midrule

Baseline
&  &  &  &  &
16.57 / 52.87 / 60.66
& 0.06 / 0.22 / 6.04
& 0.01 / 0.02 / 0.64 \\
\addlinespace[2pt]
\midrule

\multirow{3}{*}{Variants}
& \cmark &  &  &  &
17.24 / 54.59 / 62.36
& 0.06 / 0.19 / 4.47
& 0.01 / 0.02 / 0.47 \\

& \cmark & \cmark &  &  &
17.53 / 55.53 / 63.39
& 0.06 / 0.18 / 3.80
& 0.01 / 0.02 / 0.38 \\

& \cmark & \cmark & \cmark &  &
18.68 / 58.34 / 66.19
& 0.06 / 0.15 / 2.26
& 0.00 / 0.01 / 0.20 \\
\midrule

GeoMix
& \cmark & \cmark & \cmark & \cmark &
\B{21.51 / 65.61 / 73.76}
& \B{0.05 / 0.11 / 0.52}
& \B{0.00 / 0.01 / 0.05} \\
\bottomrule
\end{tabular}
}
\end{table*}

\paragraph{Generalizability.}
We evaluate generalizability along two axes. For cross-dataset transfer, the mix-trained model is directly evaluated on 7Scenes and Cambridge Landmarks without finetuning, maintaining strong accuracy despite the domain shift, as discussed above. For cross-detector transfer, we test on two detectors entirely unseen during training: R2D2~\cite{revaud2019r2d2} and DeDoDe-v2~\cite{edstedt2024dedodev2}. As shown in \cref{tab: matching_detectors}, mix-training outperforms single-training by a large margin on both, with the median rotation error dropping from $1.94^\circ$ to $0.34^\circ$ for R2D2 and from $0.39^\circ$ to $0.17^\circ$ for DeDoDe-v2, confirming that mix-training encourages reliance on geometric structure rather than detector-specific patterns.

\paragraph{Ablation Study.}
\cref{tab:abla} presents a progressive ablation on MegaDepth with $k=10$ retrieved images, starting from the A2-GNN~\cite{zhang2025a2gnnangleannulargnnvisual} baseline. We first add the learnable global context (GC) nodes, which aggregate long-range dependencies via cross-attention; this consistently improves all AUC thresholds, confirming that global aggregation helps disambiguate correspondences. We then inject directional and distance-aware embeddings into the two local geometry branches (annular convolution and maxpooling), with the maxpooling branch contributing a notably larger boost, indicating that geometry guidance is most effective at the feature aggregation stage. Finally, mix-training contributes the single largest gain, improving AUC@5/10px by over 7\%, consistent with our finding that detector diversity acts as a strong regularizer. The full model improves AUC by +4.94/+12.74/+13.10\% over the baseline and reduces the 75\%-quantile rotation error from $6.04^\circ$ to $0.52^\circ$.


\begin{figure}[!t]
  \centering
  \includegraphics[width=\linewidth]{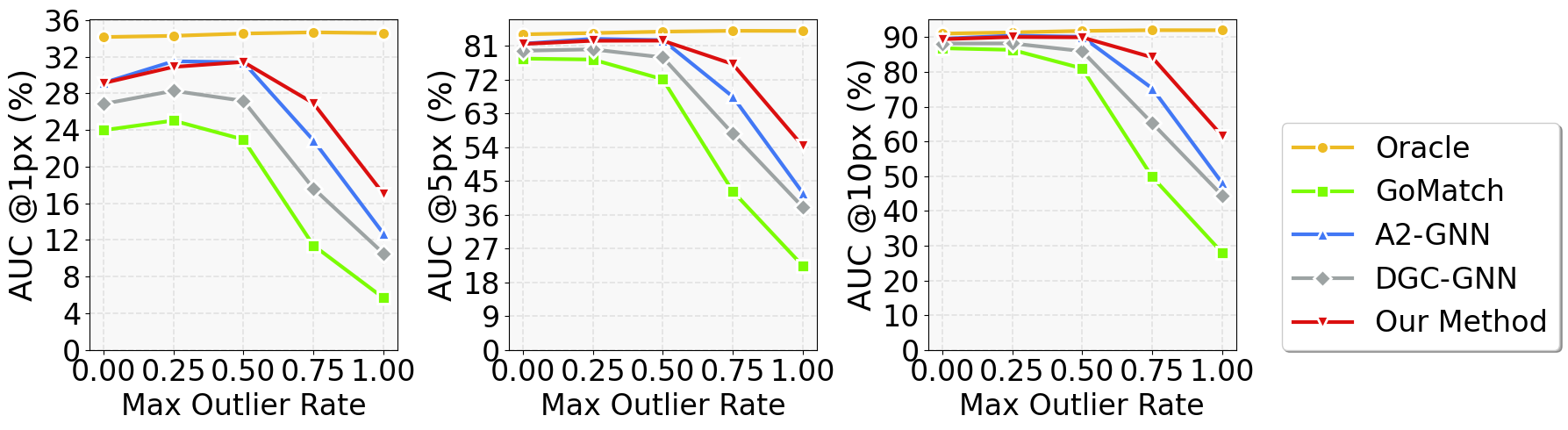}
  \caption{Sensitivity to outlier ratio. Reprojection AUC at 1, 5, and 10\,px thresholds for prior descriptor-free methods and GeoMix as the outlier ratio increases from 0 to 1. Oracle denotes the upper bound obtained with ground-truth matches.}
  \label{fig: outliers}
\end{figure}

\paragraph{Sensitivity to Keypoint Outliers.}
Following~\cite{zhou2022geometry, wang2024dgc,zhang2025a2gnnangleannulargnnvisual}, we evaluate robustness by varying the outlier ratio, defined as the fraction of keypoints without a ground-truth match among all input keypoints. At ratio 0, only ground-truth inliers are provided; at ratio 1, all detected 2D keypoints and top-$k$ retrieved 3D points are used without filtering, reflecting the realistic deployment setting. As shown in \cref{fig: outliers}, GeoMix consistently outperforms all baselines across the full range. Below 50\%, GeoMix closely approaches the Oracle upper bound. At ratio 1.0, GeoMix still achieves the best performance, demonstrating strong robustness to keypoint outliers in practical conditions.

\paragraph{Matching Efficiency.}
\cref{tab:aachen} reports the matching time of descriptor-free methods on Aachen Day-Night, excluding PnP-RANSAC. Despite the additional global context nodes and geometry injection modules, our GeoMix runs at 0.115s per image pair, comparable to A2-GNN at 0.108s and only $1.5\times$ slower than GoMatch at 0.077s, the lightest baseline, while achieving significantly higher localization accuracy. Compared to DGC-GNN, which requires 0.284s per image pair and is $2.5\times$ slower than ours, GeoMix offers both better accuracy and higher throughput, striking a favorable balance between efficiency and performance.
In terms of model size, GeoMix contains 3.50M parameters, comparable to A2-GNN (2.67M) and GoMatch (1.32M), while being notably smaller than DGC-GNN (5.65M). This confirms that the accuracy gains of GeoMix come without significant parameter overhead.
We also compare end-to-end query cost against on-the-fly descriptor matching on Aachen (top-5, 1{,}024 keypoints): GeoMix matches the query directly to the descriptor-free map in 111.3\,ms/query, whereas an on-the-fly hierarchical baseline (SP+SG matched to the 5 retrieved database images, then chained to 3D via their known associations) takes 138.2\,ms. SP+SG stays more accurate, but GeoMix offers a competitive descriptor-free alternative for storage- and privacy-sensitive deployments.

\section{Conclusion and Limitations}

We present GeoMix, a descriptor-free 2D--3D matching framework that enhances geometric discriminability at three complementary levels: locally via directional and distance-aware embeddings, globally via learnable context nodes that aggregate and redistribute scene-wide information, and at the training level via Mix-Training that exploits the detector-agnostic property unique to the descriptor-free setting to learn geometry generalizing across and even beyond seen detectors. Extensive experiments on four benchmarks show that GeoMix sets a new state of the art among descriptor-free methods, substantially narrowing the gap to descriptor-based pipelines while preserving compact storage, privacy, and zero descriptor maintenance overhead.
Despite these advances, a performance gap to descriptor-based methods remains, especially under high outlier ratios where purely geometric cues alone lack sufficient discriminability. Moreover, the 3D maps are reconstructed with a fixed detector, potentially introducing residual bias toward specific point distributions. The map still extends incrementally: new images are added with only their pose and 3D-point associations, and any descriptors for triangulating new geometry are used transiently and discarded. Future work could jointly diversify the 3D reconstruction side and incorporate lightweight appearance cues to further close the gap without compromising the benefits of the descriptor-free paradigm.

\section*{Acknowledgements}
This work was supported by the Research Council of Finland (projects 352788, 362407, 373999, 373780, 353139, 362409, 373997, 373778) and the Finnish Doctoral Program Network in Artificial Intelligence, AI-DOC (decision number VN\slash3137\slash2024-OKM-6). We acknowledge the computational resources provided by the CSC-IT Center for Science, Finland.

\newpage
%
%
\bibliographystyle{splncs04}
\bibliography{main}
\end{document}


\title{GeoMix: Supplementary Material}
\titlerunning{GeoMix: Supplementary Material}
\author{}
\institute{}
\maketitle

\appendix

\section{Implementation Details}
\label{sec:impl_details}

We provide additional architecture and training details beyond those in the main paper. The feature dimension is $d{=}128$ throughout the network. The feature encoder uses two separate ResNet-style MLPs with 12 residual layers each for bearing vectors and RGB colors, using instance normalization and ReLU activations. The GNN module consists of $L{=}3$ attention layers in a self--cross--self pattern: self-attention for local geometry aggregation, cross-attention between 2D and 3D point sets, and a final self-attention layer. The aggregation MLP $\phi_{agg}$ and broadcasting MLP $\phi_{broad}$ each consist of two linear layers with instance normalization and ReLU. For the Optimal Transport layer, we run up to $T{=}20$ Sinkhorn iterations with a regularization factor $\epsilon{=}0.1$ and a convergence threshold of $10^{-6}$; the dustbin cost is initialized to $1.0$ and learned during training. The outlier rejection threshold is $\tau{=}0.5$. During Mix-Training, each query image is paired with all three detectors, \ie, SIFT, SuperPoint, and DISK, per epoch, effectively tripling the training data. We use a maximum of 1,024 keypoints per image during training.

\paragraph{Cross-Modal Cross-Attention.}
As described in the main paper, the GNN alternates between self-attention and cross-attention layers in a self--cross--self pattern. The self-attention layers perform intra-modal local geometry aggregation and global context broadcasting via the GC nodes, producing per-node features $\mathbf{f}_i^{self}$ and updated GC nodes $\mathcal{G}''$. Here we detail the cross-modal cross-attention layer, which enables information exchange between the 2D and 3D point sets. Since this layer operates across modalities, we use superscripts ${}^{2D}$ and ${}^{3D}$ to distinguish the two sides.

The GC nodes from each modality are first concatenated with their respective spatial features to form augmented representations:
\begin{equation}
\tilde{\mathcal{F}}^{2D} = [\mathcal{F}^{self,2D} \concat \mathcal{G}''^{2D}], \quad \tilde{\mathcal{F}}^{3D} = [\mathcal{F}^{self,3D} \concat \mathcal{G}''^{3D}],
\end{equation}
where $\mathcal{F}^{self,2D}$ and $\mathcal{F}^{self,3D}$ collect all $\mathbf{f}_i^{self}$ from the preceding self-attention layer for the 2D and 3D sides respectively, and $\mathcal{G}''^{2D}$, $\mathcal{G}''^{3D}$ are the corresponding GC nodes. Cross-modal attention is then computed bidirectionally with 4 attention heads:
\begin{align}
\mathbf{f}_i^{cross,2D} &= \mathbf{f}_i^{self,2D} + \phi_{cross}\!\left([\mathbf{f}_i^{self,2D} \concat \operatorname{CA}(\tilde{\mathbf{f}}_i^{2D},\, \tilde{\mathcal{F}}^{3D})]\right), \label{eq:cross2d}\\
\mathbf{f}_j^{cross,3D} &= \mathbf{f}_j^{self,3D} + \phi_{cross}\!\left([\mathbf{f}_j^{self,3D} \concat \operatorname{CA}(\tilde{\mathbf{f}}_j^{3D},\, \tilde{\mathcal{F}}^{2D})]\right), \label{eq:cross3d}
\end{align}
where $\operatorname{CA}(\cdot)$ denotes cross-attention as in the main paper, and $\phi_{cross}$ is an MLP with the same structure as $\phi_{agg}$ and $\phi_{broad}$ ($[2d, 2d, d]$ with instance normalization and ReLU). The GC nodes participate as additional tokens alongside the spatial features, serving as information bridges that carry modality-specific global context into the cross-modal exchange. After the cross-attention, the updated GC nodes are separated from the spatial features and passed to the subsequent self-attention layer for further intra-modal refinement.

\section{Design Justification for Global Context Nodes}
\label{sec:gc_justification}

We adopt learnable global context nodes rather than full self-attention over all $N$ nodes, which incurs $O(N^2)$ cost and is prohibitive for large point sets. The $N_g$ global context nodes, where $N_g \ll N$, serve as an information bottleneck that reduces the complexity to $O(N \cdot N_g)$. The intermediate self-attention step among global context nodes is also essential: without it, each global context node aggregates and broadcasts independently, leading to redundant representations; inter-node communication encourages the nodes to capture complementary aspects of the global structure. Finally, 2D and 3D point sets maintain separate global context nodes because their geometric distributions differ fundamentally: 2D keypoints lie on the image plane while 3D points span the scene volume, and sharing nodes would conflate modality-specific patterns.

\section{Dataset Details}
\label{sec:dataset_details}

\paragraph{MegaDepth Co-Visibility Filtering.}
Following GoMatch~\cite{zhou2022geometry}, we sample up to 500 queries per scene. For each query, we collect co-visible database views that share at least 35\% visual overlap, computed as the ratio of commonly observed 3D points to the total 3D points visible in the query view. Queries with fewer than three co-visible views are discarded. After filtering, the dataset contains 18,881 training queries across 99 scenes, 3,146 validation queries across 16 scenes, and 7,344 test queries across 49 scenes.

\paragraph{Correspondence Retrieval Strategy.}
Since SIFT is the native detector used for the original SfM reconstruction in MegaDepth, it yields far more inlier correspondences per image pair, 112 on average, than SuperPoint with 33 or DISK with 47, which operate in a cross-detector setting. To compensate for the lower inlier counts of learning-based detectors, we retrieve correspondences from the top-3 database images for SuperPoint and DISK while keeping the top-1 setting for SIFT.

\paragraph{Ground Truth Correspondences.}
During training, we obtain the ground truth correspondences by reprojecting the 3D point clouds from the top-$k$ retrieved images onto the query image plane. A correspondence is considered ground truth if the reprojection error is less than 0.001 in normalized image coordinates.

\paragraph{Visual Localization Benchmarks.}
Cambridge Landmarks~\cite{kendall2015posenet} is a medium-scale outdoor dataset with SfM-based 3D reconstructions; we evaluate four of its six scenes using SuperPoint features. 7Scenes~\cite{shotton2013scene} is a small-scale indoor dataset with seven scenes captured by an RGB-D camera; we evaluate all scenes using SIFT features.

\paragraph{Aachen Day-Night Retriangulation.}
Aachen Day-Night~\cite{sattler2018benchmarking} requires localizing 922 queries, including 824 daytime and 98 nighttime, against a 3D model built from 4,328 daytime database images. The nighttime queries are particularly challenging due to drastic illumination changes. We reconstruct the 3D model using the hloc toolbox~\cite{sarlin2019coarse}, retriangulating 3D points with four detectors: SIFT, SuperPoint, R2D2, and DISK, over the same pre-computed database image pairs. During retriangulation, SIFT features are matched with nearest-neighbor ratio matching, SuperPoint features with SuperGlue~\cite{sarlin2020superglue}, and R2D2 and DISK features with mutual nearest-neighbor matching. A maximum of 2,048 keypoints are extracted per image during this stage. At evaluation time, keypoints are downsampled to 1,024 for both 2D queries and the 3D map, and matching is restricted to the top-5 retrieved database images. Since the query image is matched independently against the 3D points from each retrieved database image, we apply a retrieval-rank weighting to merge the per-image matching scores: the matching score from the $r$-th ranked database image is scaled by $e^{-0.05r}$, and when multiple database images observe the same 3D point, the highest weighted score is retained. All compared methods are evaluated on the same retriangulated 3D models to ensure a fair comparison.

\begin{table*}[!t]
\begin{center}
\caption{Ablation on the number of global context nodes $N_g$, evaluated on MegaDepth. Best results per detector are in \textbf{bold}.}
\label{tab:abla_register}
\renewcommand\arraystretch{1.2}
\setlength{\tabcolsep}{4pt}
\scriptsize
\begin{tabular}{llccc}
\toprule
\multirow{2}{*}{Detector} &
\multirow{2}{*}{$N_g$} &
  Reproj. AUC (\%) &
  Rotation ($^\circ$) &
  Translation \\

 & &
  @1 / 5 / 10px ($\uparrow$) &
  \multicolumn{2}{c}{Quantile @25 / 50 / 75\% $(\downarrow)$}
   \\ \midrule

\multirow{3}{*}{SIFT}
& 2 & 20.81 / 64.02 / 72.20 & 0.05 / 0.12 / \phantom{1}0.65 & \B{0.00} / \B{0.01} / 0.06 \\
& 4 & \B{21.51} / \B{65.61} / \B{73.76} & \B{0.05} / \B{0.11} / \phantom{1}\B{0.52} & \B{0.00} / \B{0.01} / \B{0.05} \\
& 8 & 21.39 / 65.31 / 73.37 & \B{0.05} / \B{0.11} / \phantom{1}0.54 & \B{0.00} / \B{0.01} / \B{0.05} \\
\midrule
\multirow{3}{*}{SuperPoint}
& 2 & \phantom{1}0.11 / 20.45 / 36.78 & 0.62 / 1.60 / \phantom{1}7.41 & \B{0.05} / \B{0.14} / 0.80 \\
& 4 & \phantom{1}\B{0.15} / 20.84 / 36.89 & \B{0.60} / 1.57 / \phantom{1}7.57 & \B{0.05} / \B{0.14} / 0.77 \\
& 8 & \phantom{1}0.12 / \B{21.12} / \B{37.25} & \B{0.60} / \B{1.54} / \phantom{1}\B{7.05} & \B{0.05} / \B{0.14} / \B{0.73} \\
\midrule
\multirow{3}{*}{DISK}
& 2 & \phantom{1}\B{0.04} / 30.10 / 48.60 & \B{0.33} / 0.97 / \phantom{1}4.35 & \B{0.03} / 0.09 / 0.41 \\
& 4 & \phantom{1}\B{0.04} / 30.30 / 48.80 & \B{0.33} / 0.96 / \phantom{1}4.22 & \B{0.03} / \B{0.08} / 0.41 \\
& 8 & \phantom{1}\B{0.04} / \B{30.50} / \B{49.10} & \B{0.33} / \B{0.93} / \phantom{1}\B{3.97} & \B{0.03} / \B{0.08} / \B{0.37} \\
\bottomrule
\end{tabular}
\end{center}
\end{table*}

\section{Additional Ablation Studies}
\label{sec:abla_additional}

We conduct additional ablation experiments on MegaDepth to study the effect of two factors: the number of global context nodes and the training dataset combination. All models are evaluated with three keypoint detectors: SIFT, SuperPoint, and DISK.

\begin{table*}[!t]
\begin{center}
\caption{Ablation on training dataset combination, evaluated on MegaDepth. Best results per detector are in \textbf{bold}.}
\label{tab:abla_dataset}
\renewcommand\arraystretch{1.2}
\setlength{\tabcolsep}{4pt}
\scriptsize
\begin{tabular}{llccc}
\toprule
\multirow{2}{*}{Detector} &
\multirow{2}{*}{Dataset} &
  Reproj. AUC (\%) &
  Rotation ($^\circ$) &
  Translation \\

 & &
  @1 / 5 / 10px ($\uparrow$) &
  \multicolumn{2}{c}{Quantile @25 / 50 / 75\% $(\downarrow)$}
   \\ \midrule

\multirow{3}{*}{SIFT}
& SIFT+SP & 20.42 / 62.86 / 70.98 & \B{0.05} / 0.12 / \phantom{1}0.83 & \B{0.00} / \B{0.01} / 0.08 \\
& SIFT+DISK & 20.74 / 64.09 / 72.18 & \B{0.05} / 0.12 / \phantom{1}0.64 & \B{0.00} / \B{0.01} / 0.06 \\
& SIFT+SP+DISK & \B{21.51} / \B{65.61} / \B{73.76} & \B{0.05} / \B{0.11} / \phantom{1}\B{0.52} & \B{0.00} / \B{0.01} / \B{0.05} \\
\midrule
\multirow{3}{*}{SuperPoint}
& SIFT+SP & \phantom{1}0.10 / 19.33 / 34.73 & 0.64 / 1.81 / \phantom{1}9.06 & 0.06 / 0.16 / 0.97 \\
& SIFT+DISK & \phantom{1}0.09 / 17.39 / 32.33 & 0.72 / 2.06 / 10.15 & 0.06 / 0.19 / 1.10 \\
& SIFT+SP+DISK & \phantom{1}\B{0.15} / \B{20.84} / \B{36.89} & \B{0.60} / \B{1.57} / \phantom{1}\B{7.57} & \B{0.05} / \B{0.14} / \B{0.77} \\
\midrule
\multirow{3}{*}{DISK}
& SIFT+SP & \phantom{1}0.03 / 26.70 / 44.00 & 0.41 / 1.28 / \phantom{1}6.86 & \B{0.03} / 0.11 / 0.71 \\
& SIFT+DISK & \phantom{1}\B{0.04} / 29.90 / 48.30 & 0.34 / 0.98 / \phantom{1}4.27 & \B{0.03} / \B{0.08} / \B{0.41} \\
& SIFT+SP+DISK & \phantom{1}\B{0.04} / \B{30.30} / \B{48.80} & \B{0.33} / \B{0.96} / \phantom{1}\B{4.22} & \B{0.03} / \B{0.08} / \B{0.41} \\
\bottomrule
\end{tabular}
\end{center}
\end{table*}

\paragraph{Number of Global Context Nodes.}
\cref{tab:abla_register} compares models trained with $N_g \in \{2, 4, 8\}$ global context nodes. With only $N_g{=}2$ nodes, performance drops noticeably across all detectors, \eg, SIFT AUC@10px falls from 73.76\% to 72.20\% and the 75th-percentile rotation error rises from $0.52^\circ$ to $0.65^\circ$, suggesting that two nodes provide insufficient capacity to capture the diverse global structure of the scene. Increasing from $N_g{=}4$ to $N_g{=}8$ yields comparable overall performance: SuperPoint and DISK see slight pose improvements with $N_g{=}8$ (\eg, DISK 75th-percentile rotation error: $3.97^\circ$ \vs $4.22^\circ$), while SIFT favors $N_g{=}4$ on all metrics. Given the similar accuracy and lower model complexity, we adopt $N_g{=}4$ as the default.

\paragraph{Training Dataset Combination.}
\cref{tab:abla_dataset} compares three mix-training dataset combinations: SIFT+\allowbreak SuperPoint, SIFT+\allowbreak DISK, and all three detectors combined. Two clear trends emerge. First, each two-detector subset is biased toward its included learning-based detector: SIFT+\allowbreak SuperPoint achieves 34.73\% AUC@10px on SuperPoint but only 32.33\% under SIFT+\allowbreak DISK, while conversely SIFT+\allowbreak DISK reaches 48.30\% on DISK versus 44.00\% under SIFT+\allowbreak SuperPoint. This confirms that training with a specific detector specializes the model to that detector's keypoint distribution at the cost of generalization. Second, the full three-detector combination consistently matches or surpasses both subsets across all detectors. The gain is most pronounced on SIFT, where AUC@10px improves from 70.98\%\,/\,72.18\% to 73.76\% and the 75th\nobreakdash-percentile rotation error drops from $0.83^\circ$\,/\,$0.64^\circ$ to $0.52^\circ$. This suggests that exposing the model to diverse keypoint distributions acts as a geometric regularizer, preventing overfitting to any single detector and strengthening shared geometric reasoning across all detectors.

\begin{table}[!t]
\begin{center}
\caption{SIFT-bias ablation on MegaDepth ($k{=}10$). As the 3D map is reconstructed with SIFT, we verify that the mix-training gains are not an artifact of reusing the map detector on the query side. Mix-training \emph{without} SIFT (SuperPoint+DISK) still beats SIFT-only single-training on every test detector and approaches the full three-detector mix. Best per detector in \textbf{bold}.}
\label{tab:abla_nosift}
\renewcommand\arraystretch{1.2}
\setlength{\tabcolsep}{4pt}
\scriptsize
\begin{tabular}{llc}
\toprule
Test detector & Training detectors & Reproj. AUC @1 / 5 / 10px ($\uparrow$) \\
\midrule
\multirow{3}{*}{SIFT}
& SIFT (single)       & 18.68 / 58.34 / 66.19 \\
& SP+DISK (no SIFT)   & 19.26 / 61.05 / 69.48 \\
& SIFT+SP+DISK (full) & \B{21.51} / \B{65.61} / \B{73.76} \\
\midrule
\multirow{3}{*}{SuperPoint}
& SIFT (single)       & \phantom{1}0.04 / \phantom{1}8.55 / 17.50 \\
& SP+DISK (no SIFT)   & \phantom{1}0.08 / 17.80 / 32.60 \\
& SIFT+SP+DISK (full) & \phantom{1}\B{0.15} / \B{20.84} / \B{36.89} \\
\midrule
\multirow{3}{*}{DISK}
& SIFT (single)       & \phantom{1}0.02 / 17.70 / 32.00 \\
& SP+DISK (no SIFT)   & \phantom{1}0.04 / 27.50 / 45.20 \\
& SIFT+SP+DISK (full) & \phantom{1}\B{0.04} / \B{30.30} / \B{48.80} \\
\bottomrule
\end{tabular}
\end{center}
\end{table}

\begin{table}[!t]
\begin{center}
\caption{Matcher-only metrics on MegaDepth ($k{=}10$), computed against ground-truth matches independently of PnP-RANSAC. Precision (= inlier ratio), recall, and F1 in \%. Best in \textbf{bold}.}
\label{tab:matcher_metrics}
\renewcommand\arraystretch{1.2}
\setlength{\tabcolsep}{6pt}
\scriptsize
\begin{tabular}{lccc}
\toprule
Method & Precision ($\uparrow$) & Recall ($\uparrow$) & F1 ($\uparrow$) \\
\midrule
GoMatch~\cite{zhou2022geometry} & 0.97 & 1.37 & 1.09 \\
DGC-GNN~\cite{wang2024dgc} & 2.17 & 3.50 & 2.58 \\
A2-GNN~\cite{zhang2025a2gnnangleannulargnnvisual} & 2.62 & 4.84 & 3.27 \\
GeoMix (Ours) & \B{4.30} & \B{8.06} & \B{5.40} \\
\bottomrule
\end{tabular}
\end{center}
\end{table}

\paragraph{Matcher-Only Metrics.}
Localization accuracy couples matcher quality with the PnP-RANSAC pose estimator. To isolate matcher quality, \cref{tab:matcher_metrics} reports precision (inlier ratio), recall, and F1 on MegaDepth ($k{=}10$), computed directly against the ground-truth match matrix. GeoMix achieves the best precision, recall, and F1, consistent with the pose-based ranking. The absolute values are low because descriptor-free matching recovers only a small fraction of the dense ground-truth correspondences, which PnP-RANSAC then tolerates.

\paragraph{Excluding SIFT from the Detector Pool.}
Because the MegaDepth 3D maps are reconstructed with SIFT, one may worry that the mix-training gains simply come from the query side reusing the map detector. \cref{tab:abla_nosift} rules this out: mix-training with only SuperPoint and DISK (no SIFT in the queries) still improves AUC@10px over SIFT-only single-training on every test detector, including SIFT itself (66.19 to 69.48), and recovers most of the gain of the full three-detector mix. The improvement therefore stems from detector diversity rather than a SIFT-induced bias.

\paragraph{Effect of Outlier Rejection.}
The Sinkhorn layer already yields soft assignment scores, but these normalized couplings remain unreliable under the high outlier ratios typical of descriptor-free matching. The learning-based outlier rejection module instead predicts an explicit per-correspondence inlier probability that is thresholded before PnP-RANSAC. Removing this module and relying solely on the Sinkhorn assignment lowers MegaDepth AUC@10px from 73.76 to 56.49, confirming that an explicit inlier classifier is essential for accurate pose estimation.

\section{Qualitative Visualization}
\label{sec:qualitative}

We provide qualitative comparisons on three scenes from Cambridge Landmarks: OldHospital, ShopFacade, and StMarysChurch. For each scene, we first show the 3D point cloud reconstructed by SfM, followed by the 2D reprojection results of four methods: GoMatch, DGC-GNN, A2-GNN, and our GeoMix. Cyan dots indicate 2D keypoints detected in the query image, green dots indicate inlier matches, and red dots indicate outlier matches. Our method consistently produces more inliers and fewer outliers across different scenes.

\begin{figure}[!t]
\centering
\includegraphics[width=0.65\linewidth,trim=400 150 400 100,clip]{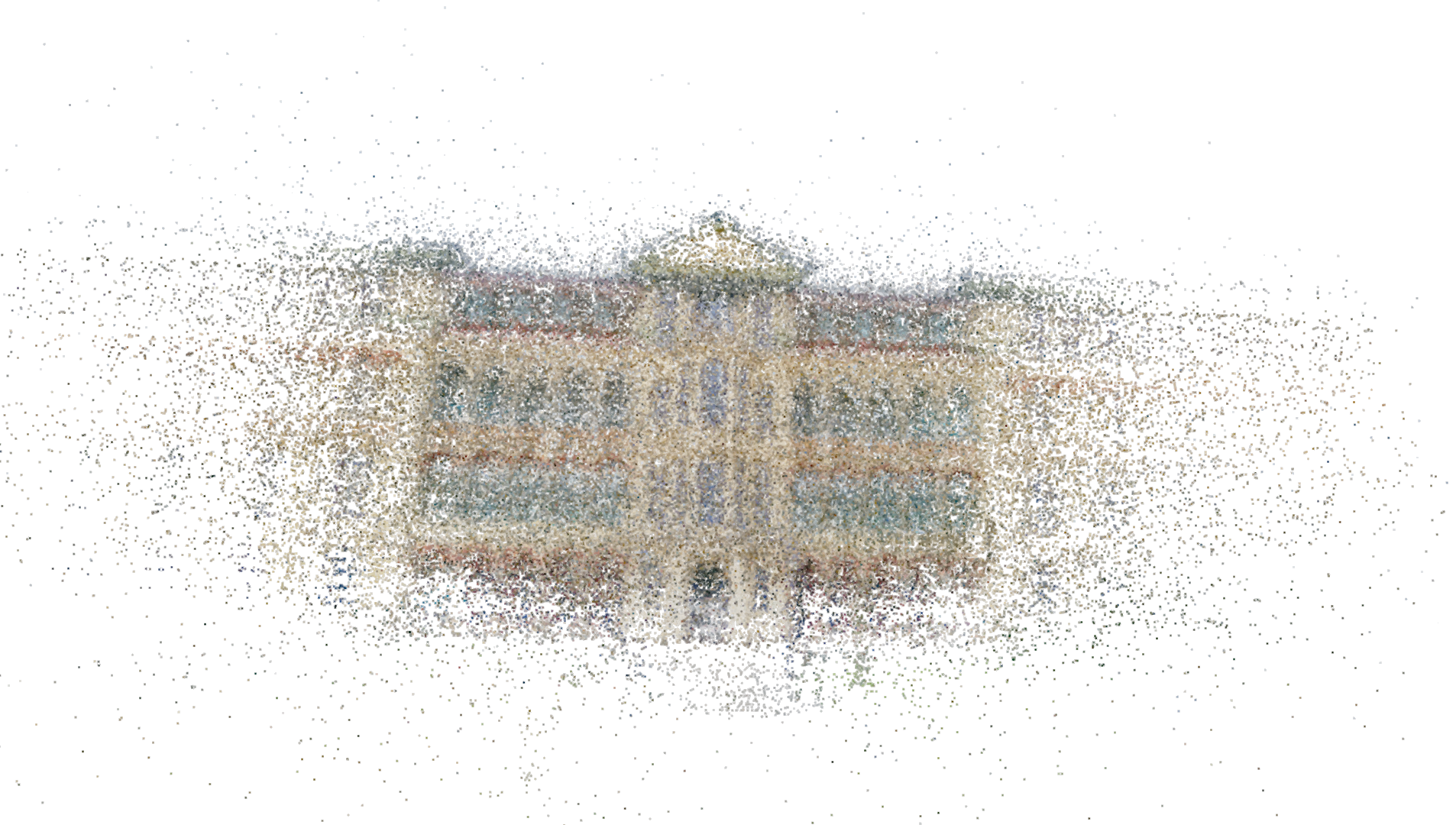}\\[4pt]
\begin{tabular}{cc}
\includegraphics[width=0.48\linewidth]{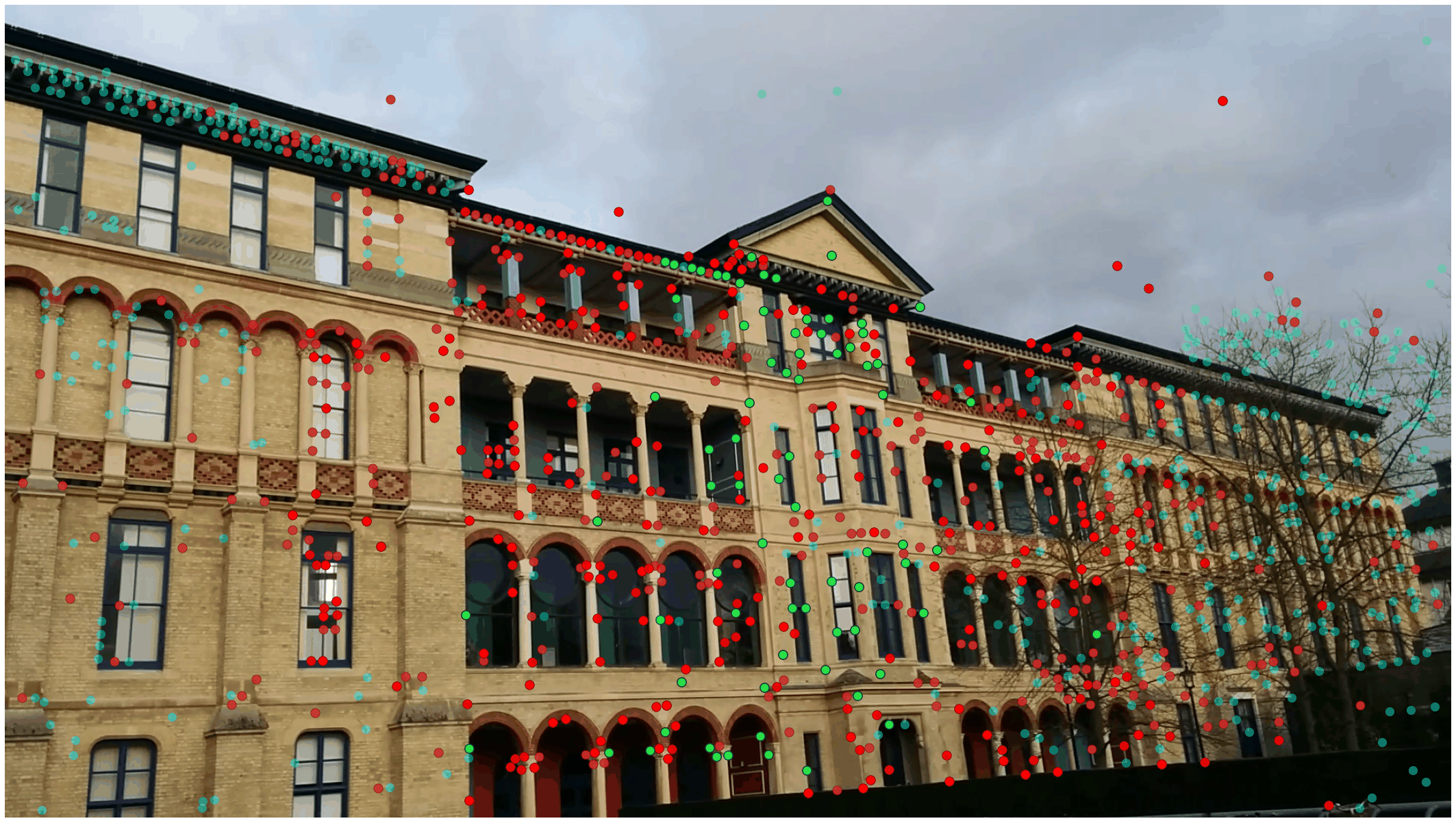} &
\includegraphics[width=0.48\linewidth]{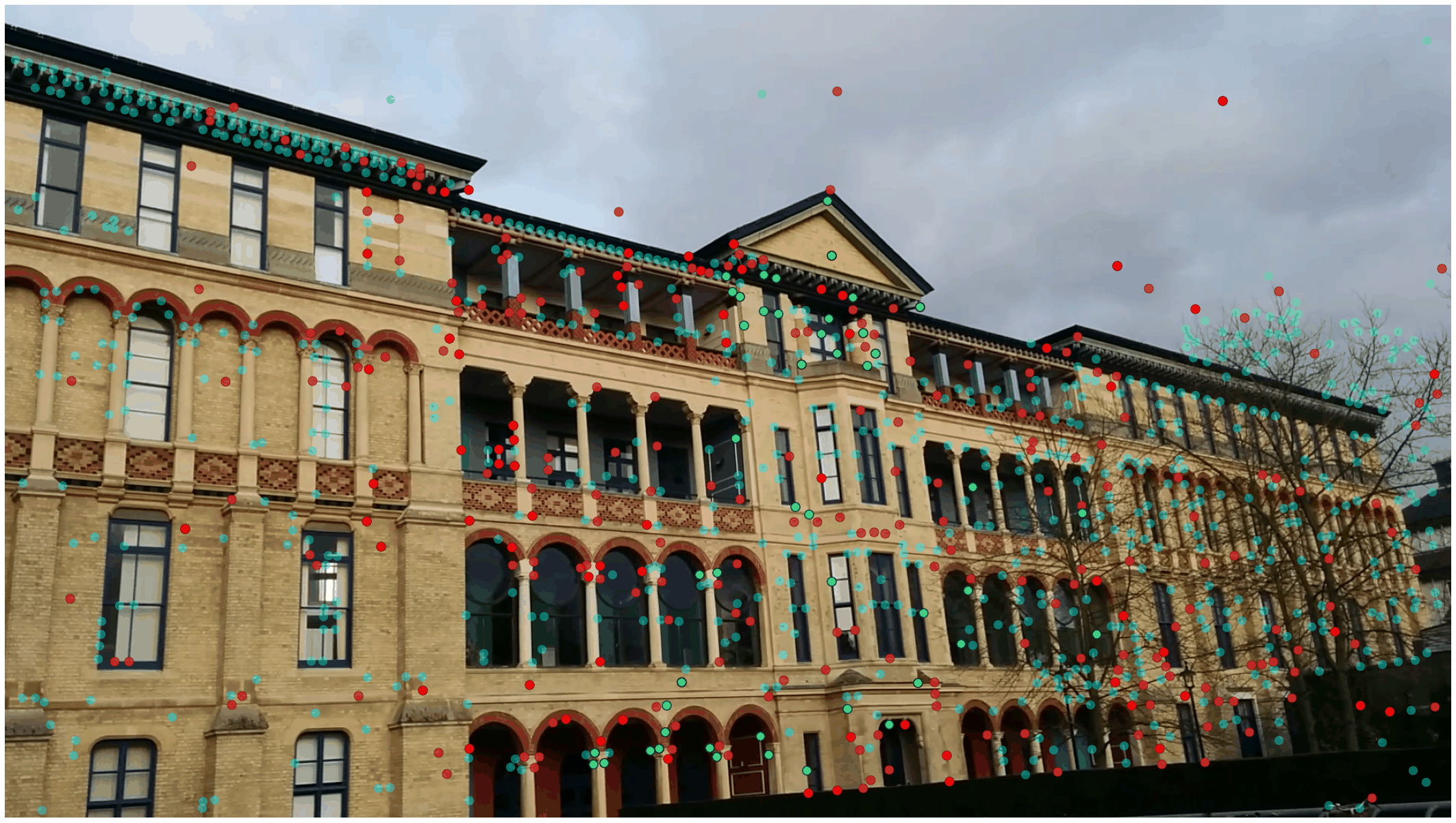} \\
(a) GoMatch & (b) DGC-GNN \\[4pt]
\includegraphics[width=0.48\linewidth]{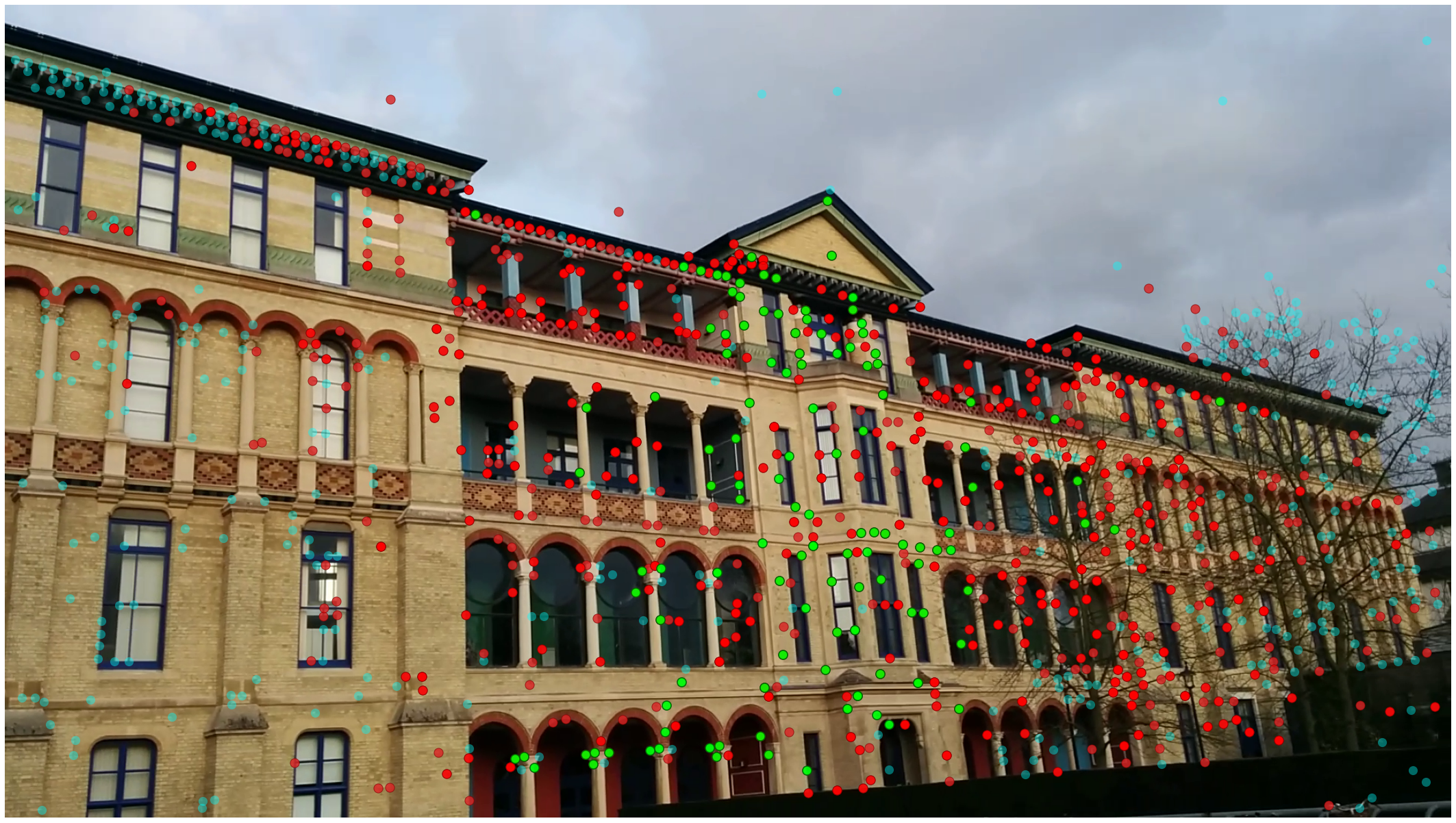} &
\includegraphics[width=0.48\linewidth]{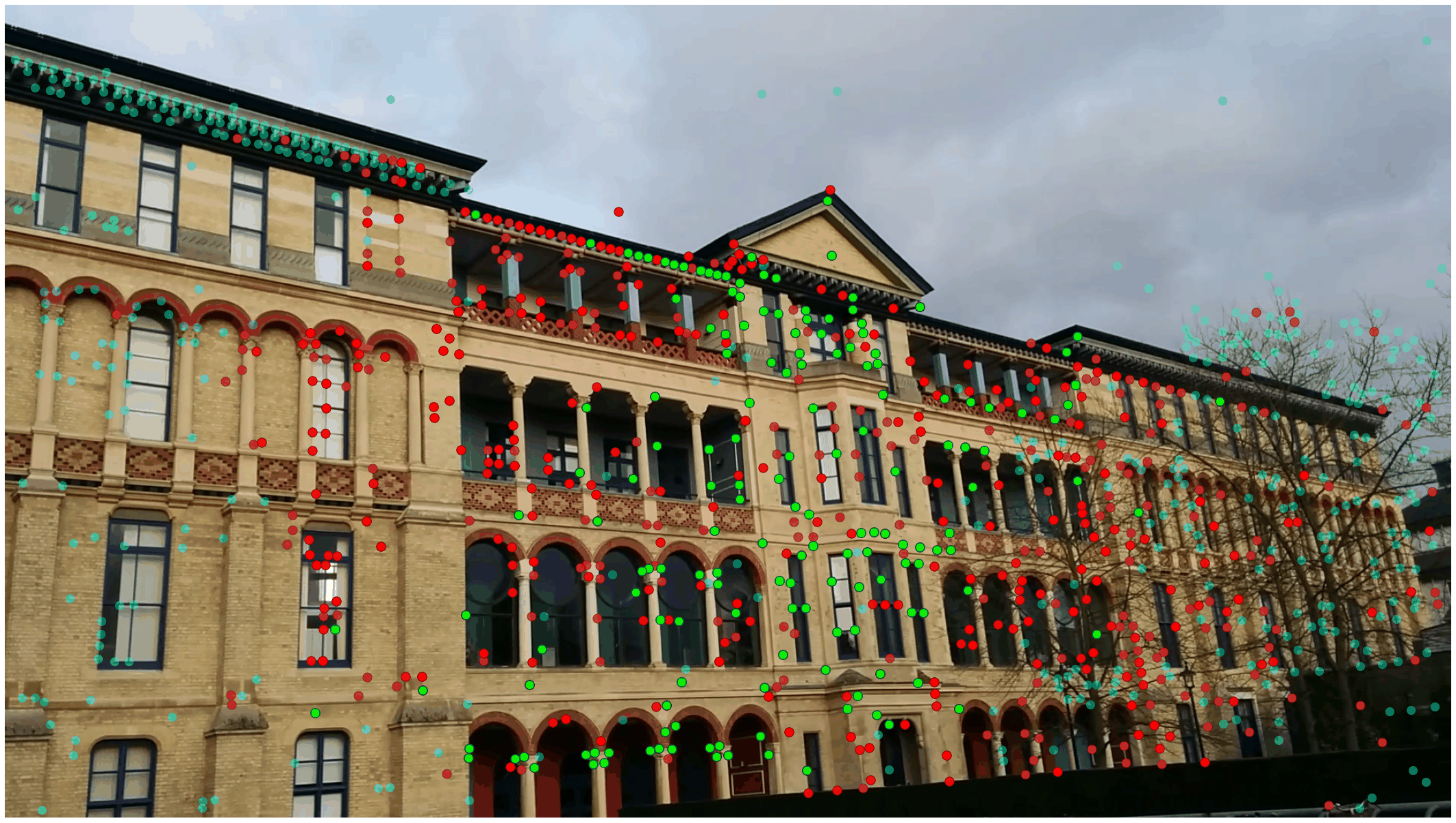} \\
(c) A2-GNN & (d) GeoMix (Ours) \\
\end{tabular}
\caption{Qualitative comparison on the OldHospital scene. Top: 3D point cloud. Bottom: 2D reprojection results. \textcolor{cyan}{Cyan}: 2D keypoints; \textcolor{green}{Green}: inlier matches; \textcolor{red}{Red}: outlier matches.}
\label{fig:vis_oldhospital}
\end{figure}

\begin{figure}[!t]
\centering
\includegraphics[width=0.65\linewidth,trim=350 150 350 150,clip]{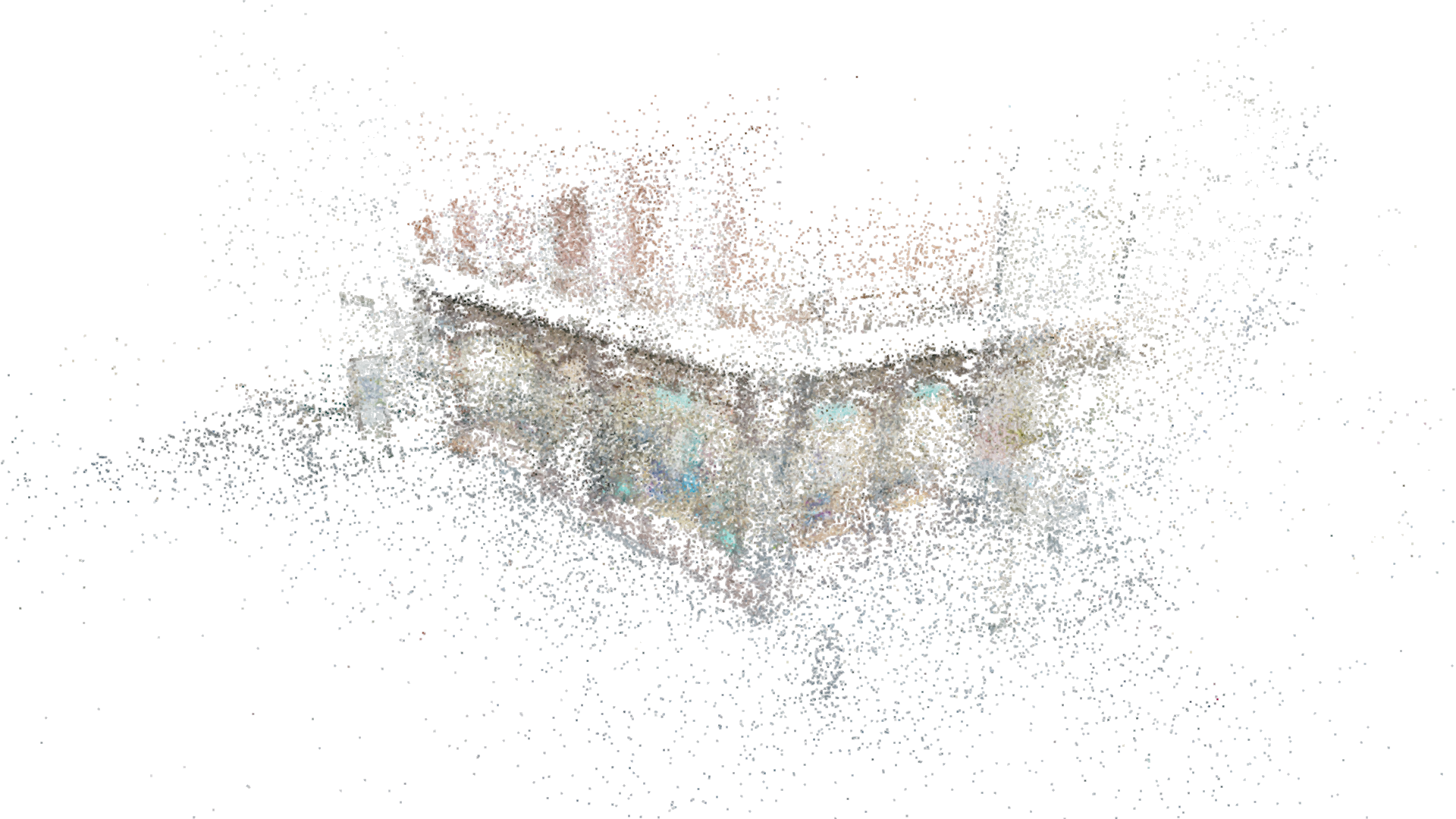}\\[4pt]
\begin{tabular}{cc}
\includegraphics[width=0.48\linewidth]{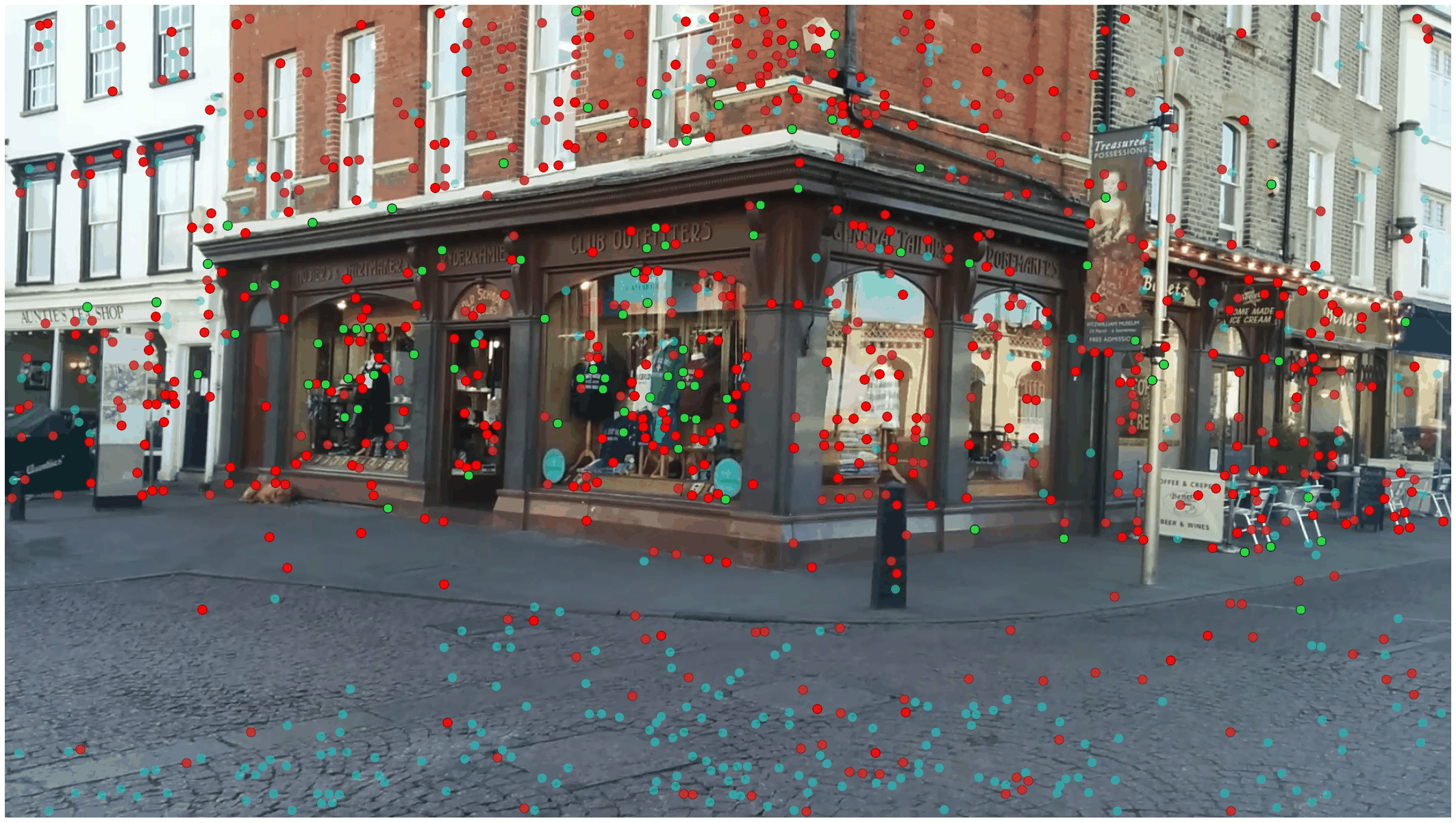} &
\includegraphics[width=0.48\linewidth]{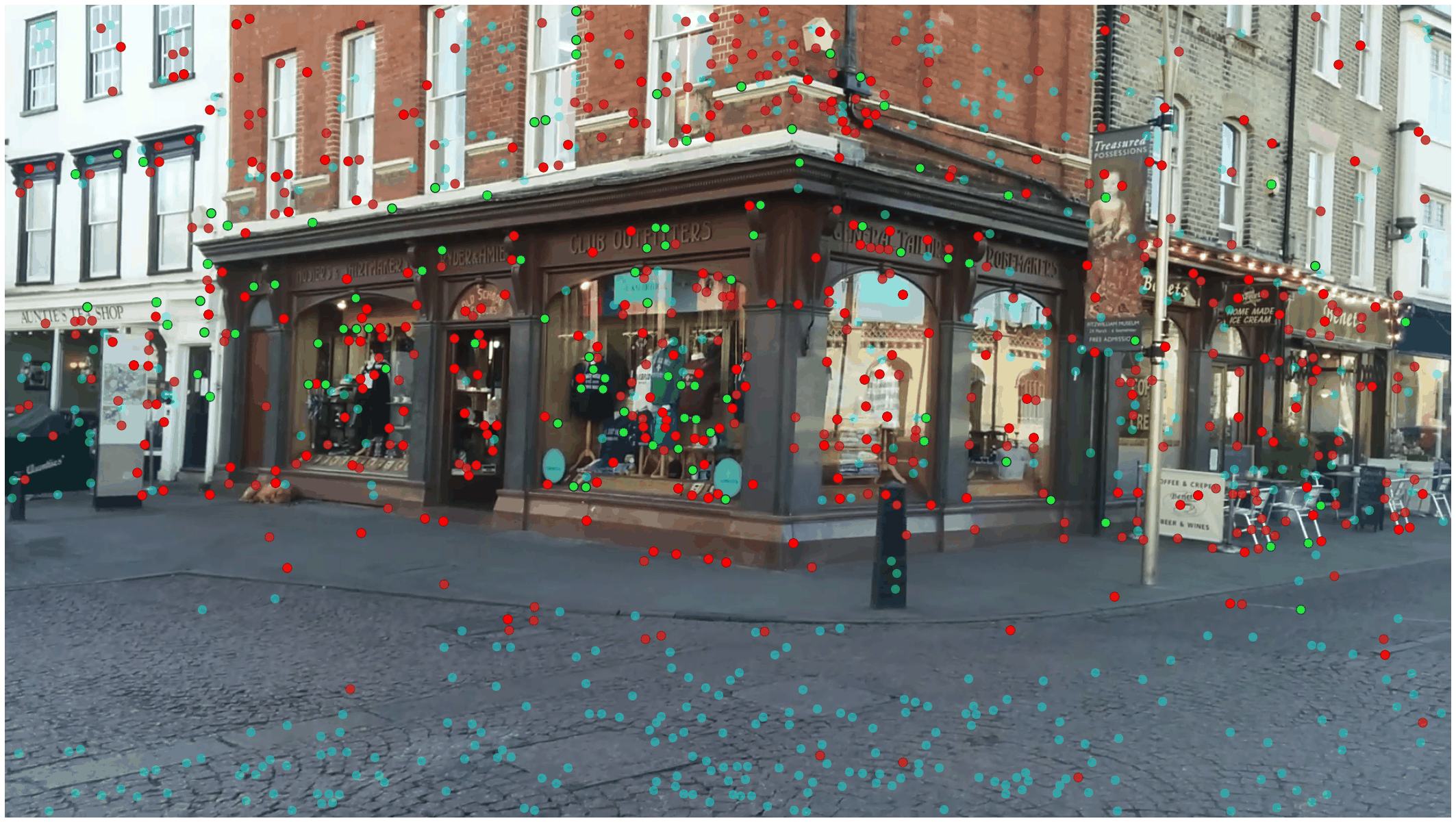} \\
(a) GoMatch & (b) DGC-GNN \\[4pt]
\includegraphics[width=0.48\linewidth]{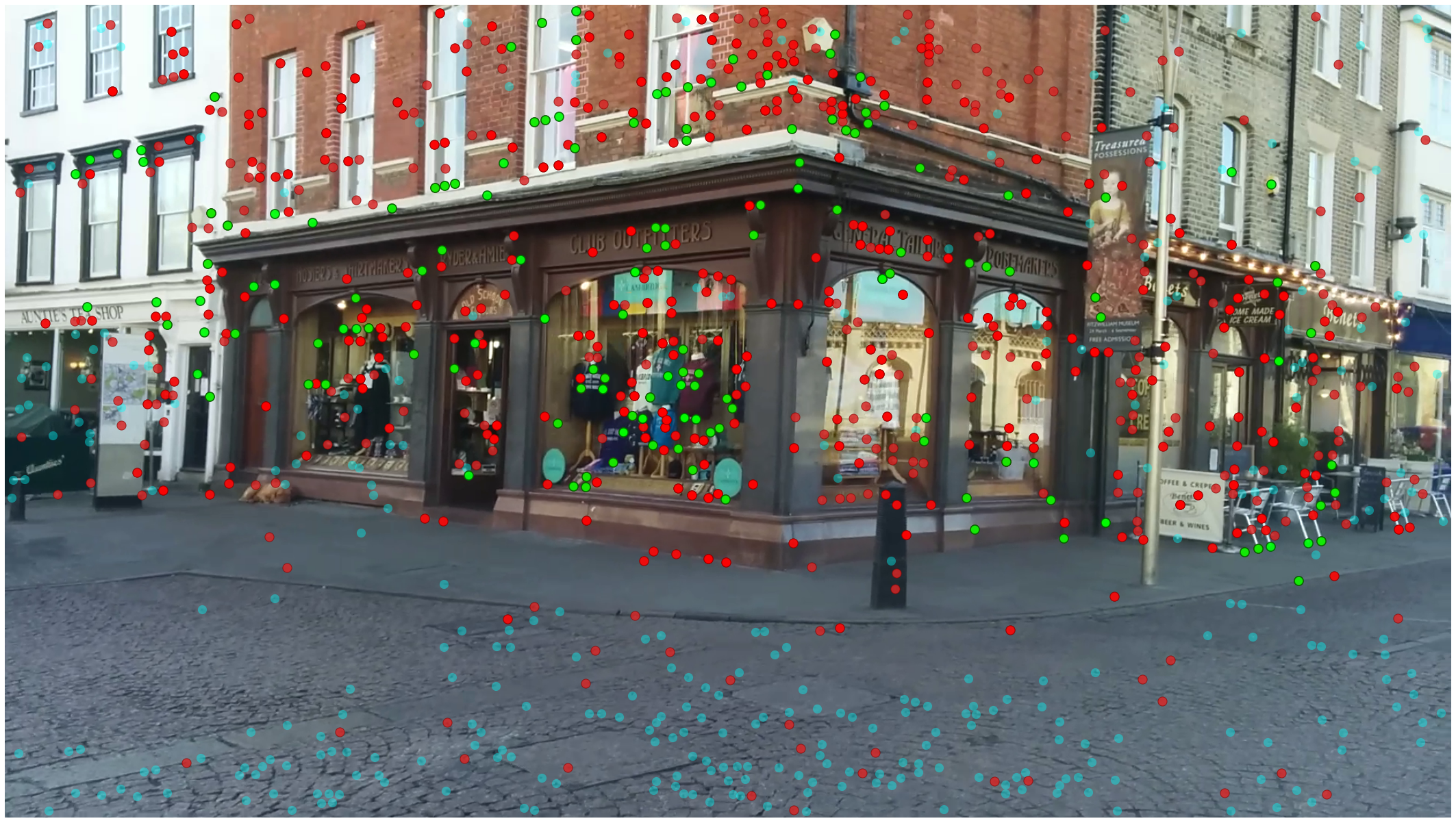} &
\includegraphics[width=0.48\linewidth]{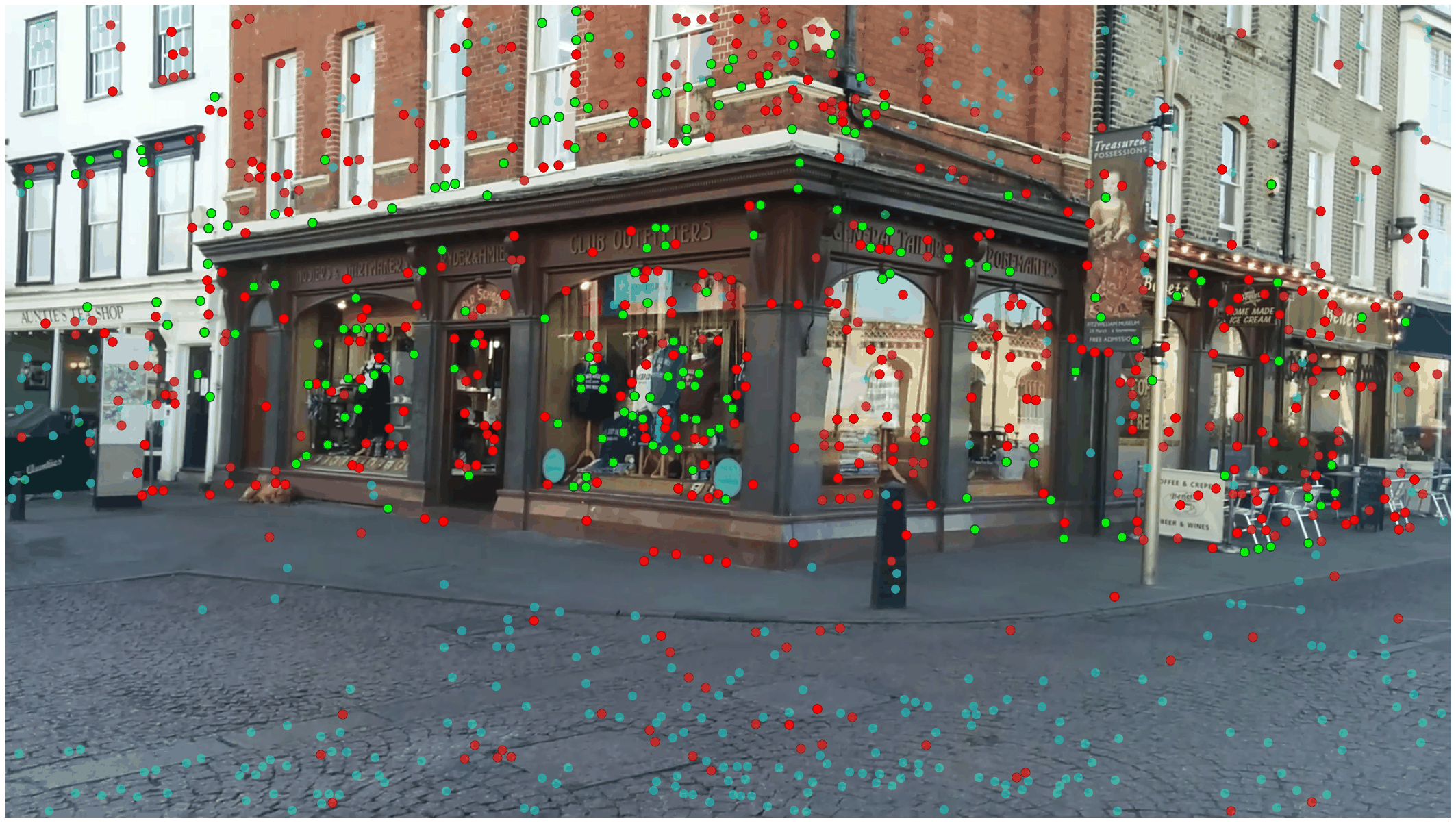} \\
(c) A2-GNN & (d) GeoMix (Ours) \\
\end{tabular}
\caption{Qualitative comparison on the ShopFacade scene. Top: 3D point cloud. Bottom: 2D reprojection results. \textcolor{cyan}{Cyan}: 2D keypoints; \textcolor{green}{Green}: inlier matches; \textcolor{red}{Red}: outlier matches.}
\label{fig:vis_shopfacade}
\end{figure}

\begin{figure}[!t]
\centering
\includegraphics[width=0.65\linewidth,trim=450 50 450 100,clip]{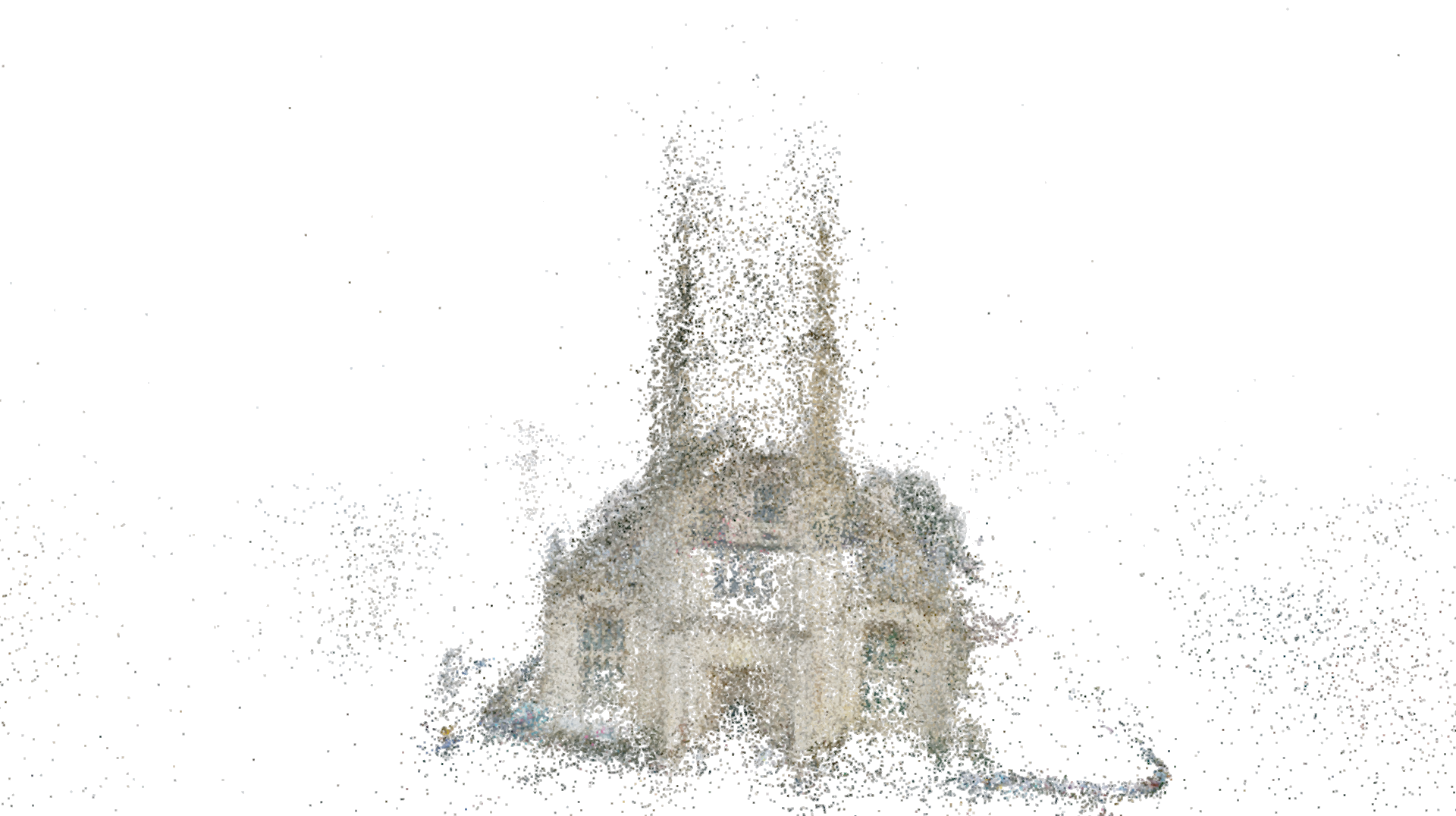}\\[4pt]
\begin{tabular}{cc}
\includegraphics[width=0.48\linewidth]{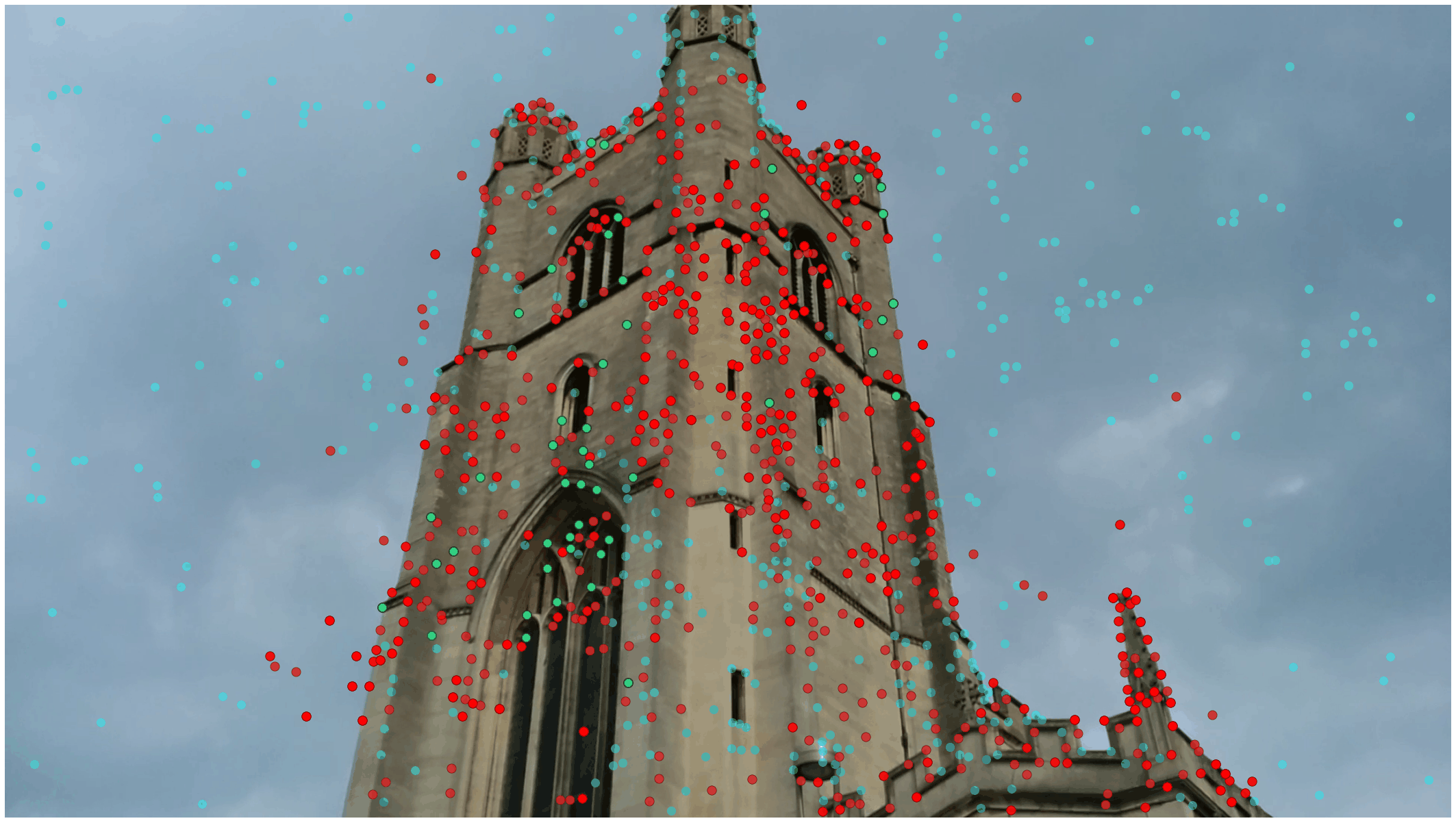} &
\includegraphics[width=0.48\linewidth]{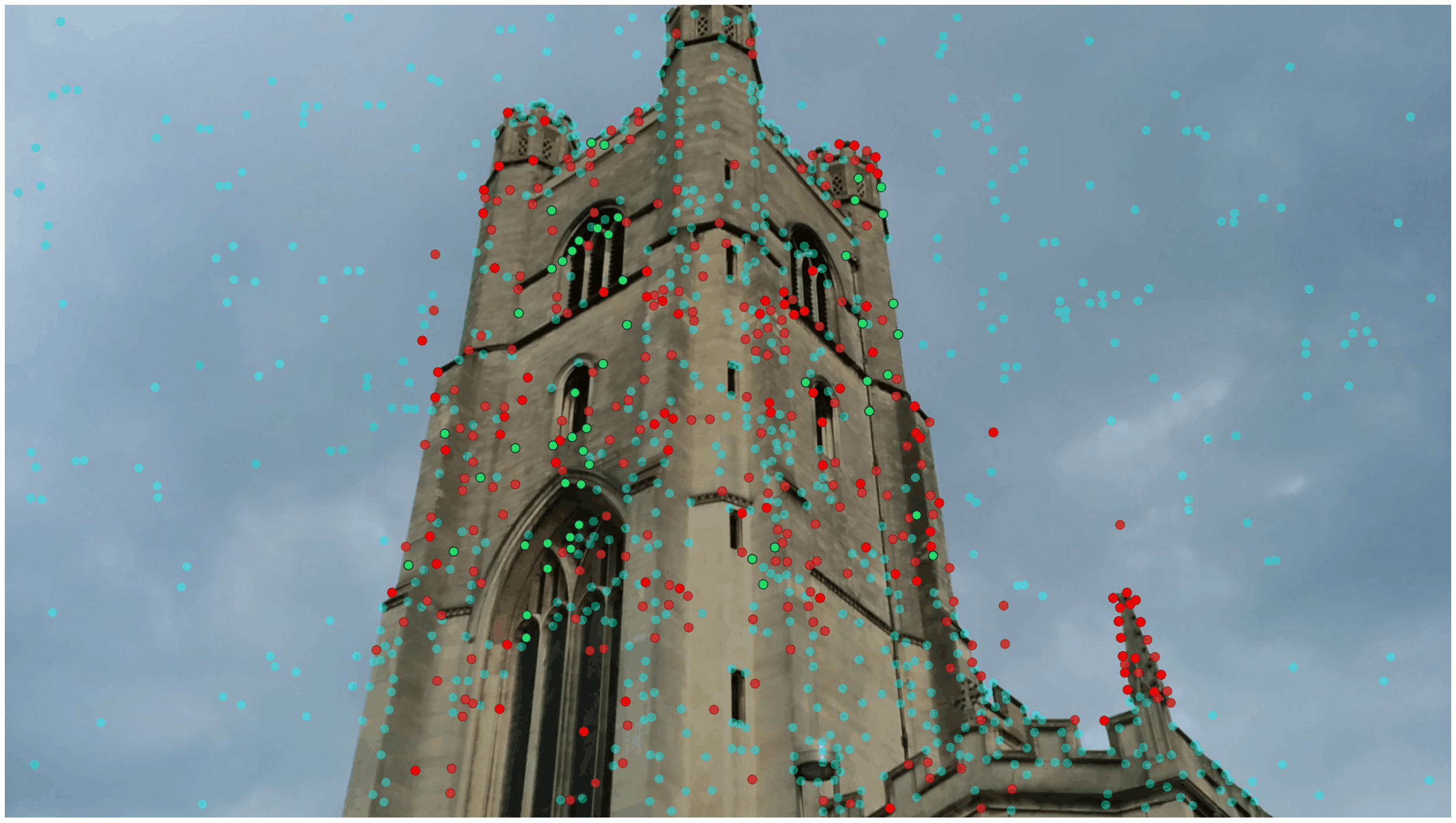} \\
(a) GoMatch & (b) DGC-GNN \\[4pt]
\includegraphics[width=0.48\linewidth]{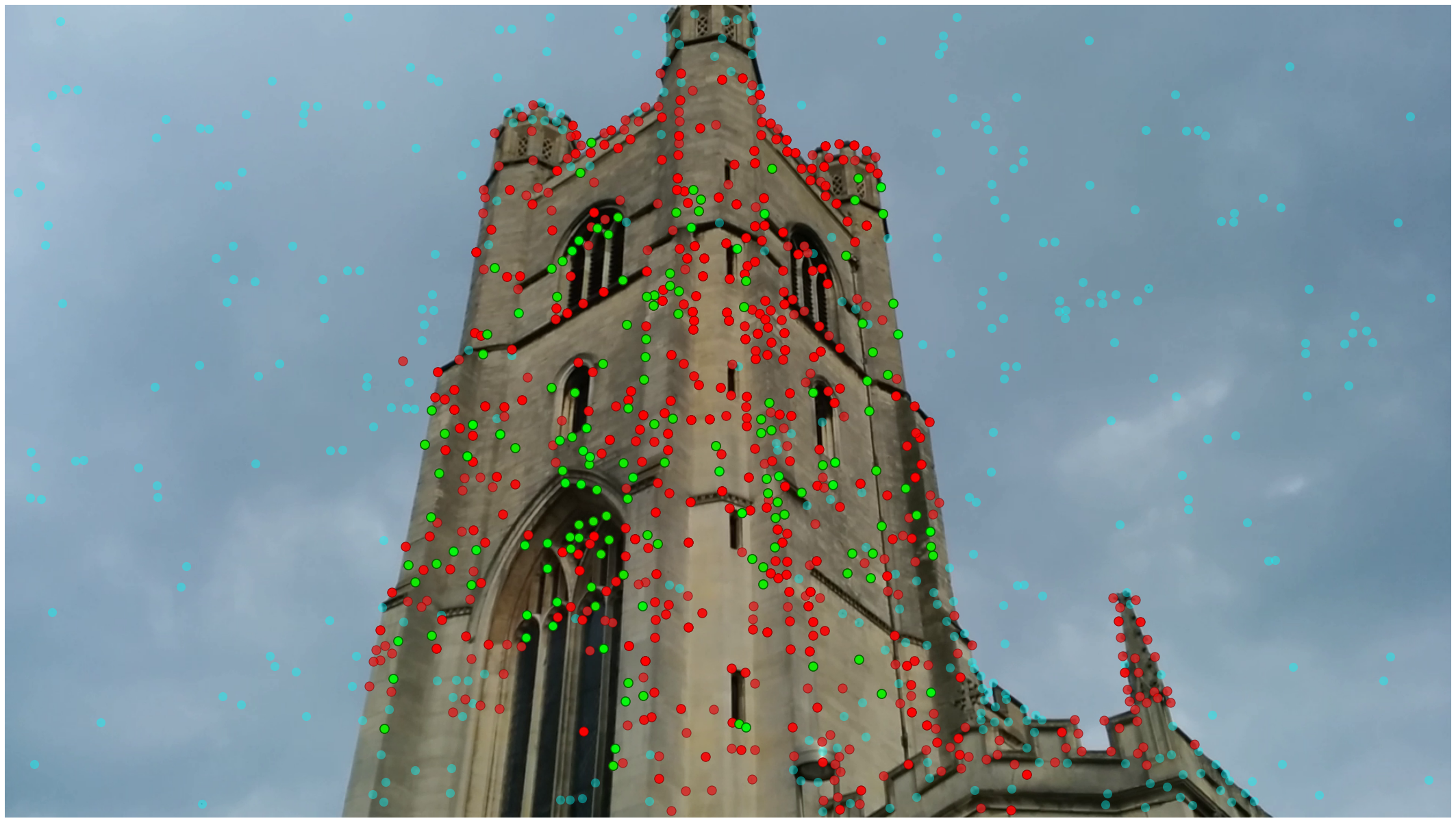} &
\includegraphics[width=0.48\linewidth]{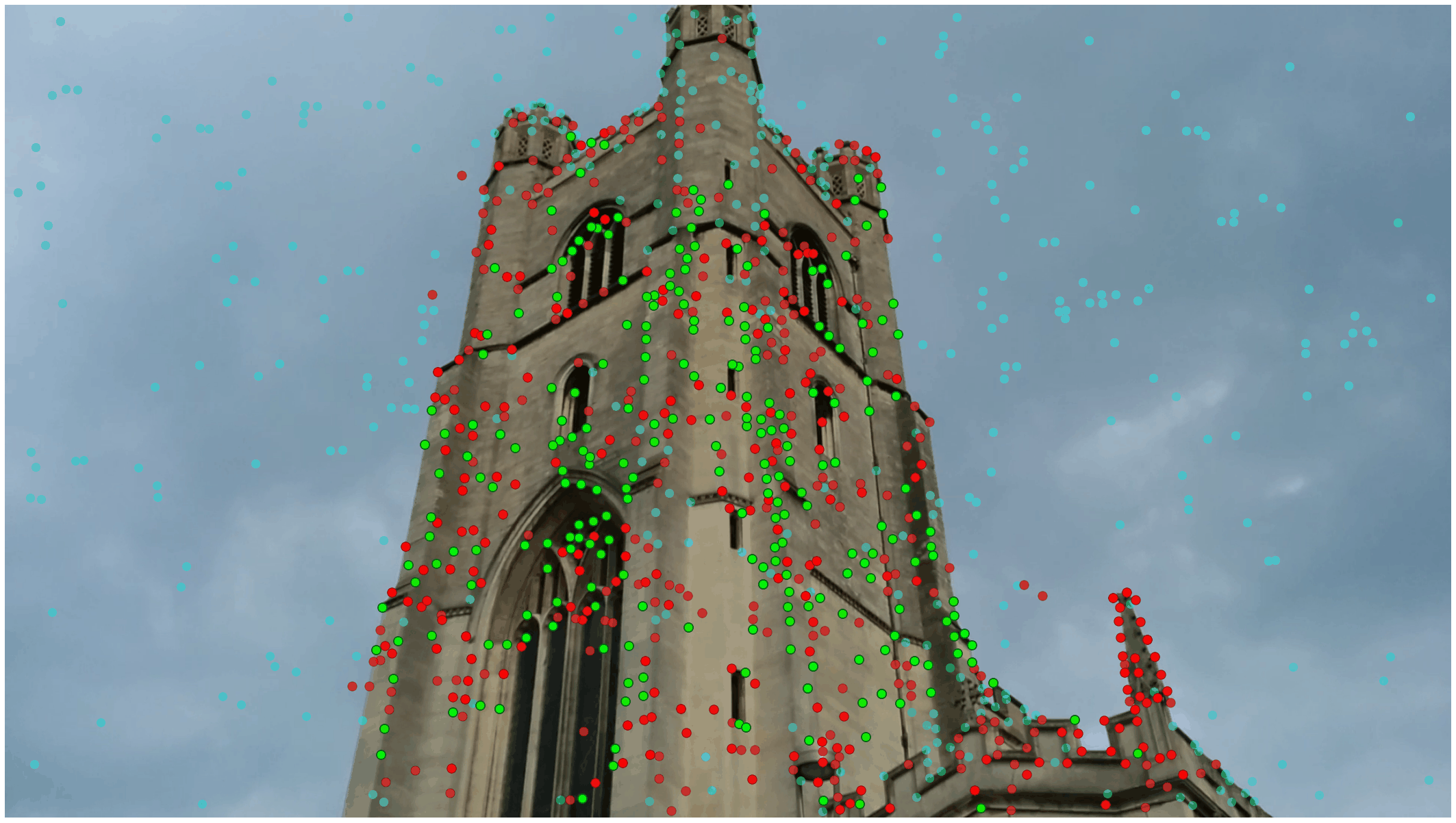} \\
(c) A2-GNN & (d) GeoMix (Ours) \\
\end{tabular}
\caption{Qualitative comparison on the StMarysChurch scene. Top: 3D point cloud. Bottom: 2D reprojection results. \textcolor{cyan}{Cyan}: 2D keypoints; \textcolor{green}{Green}: inlier matches; \textcolor{red}{Red}: outlier matches.}
\label{fig:vis_stmaryschurch}
\end{figure}

\clearpage
\bibliographystyle{splncs04}
\bibliography{main}